\documentclass[lettersize,journal]{IEEEtran}
\usepackage{amsmath,amsfonts}
\usepackage{algorithmic}
\usepackage{algorithm}
\usepackage{array}
\usepackage[caption=false,font=normalsize,labelfont=sf,textfont=sf]{subfig}
\usepackage{textcomp}
\usepackage{stfloats}
\usepackage{url}
\usepackage{verbatim}
\usepackage{graphicx}
\usepackage{cite}
\usepackage{booktabs}
\usepackage{makecell}
\usepackage{multirow}
\usepackage[colorlinks=true, linkcolor=blue, citecolor=blue]{hyperref}
\usepackage{array}
\usepackage{amssymb}
\usepackage{soul}
\begin{document}

\title{Machine Unlearning on Pre-trained Models by Residual Feature Alignment Using LoRA}

\author{Laiqiao Qin,
\IEEEmembership{Student Member, IEEE}, 
Tianqing Zhu*, \IEEEmembership{Member, IEEE},
Linlin Wang, \IEEEmembership{Student Member, IEEE},\\
and Wanlei Zhou, \IEEEmembership{Senior Member, IEEE}
\thanks{*Tianqing Zhu is the corresponding author.}
\thanks{Laiqiao Qin, Tianqing Zhu, Linlin Wang, and Wanlei Zhou are with the City University of Macau,
Macau (e-mail: isqlq@outlook.com; tianqing.zhu@uts.edu.au; linlinwang.cityu@gmail.com; wlzhou@cityu.edu.mo).}
}

\markboth{Journal of \LaTeX\ Class Files,~Vol.~14, No.~8, August~2021}%
{Shell \MakeLowercase{\textit{et al.}}: A Sample Article Using IEEEtran.cls for IEEE Journals}

\IEEEpubid{0000--0000/00\$00.00~\copyright~2021 IEEE}

\maketitle

\begin{abstract}
Machine unlearning is an emerging technology that removes a subset of the training data from a trained model without significantly affecting the model performance on the remaining data. This topic is becoming increasingly important in protecting user privacy and eliminating harmful or outdated data. The key challenge lies in effectively and efficiently unlearning specific information without compromising the model's utility on the retained data. For pre-trained models, fine-tuning is an important way to achieve the unlearning target. Previous work typically fine-tuned the entire model's parameters, which incurred significant computational costs. In addition, the fine-tuning process may cause shifts in the intermediate layer features, affecting the model's overall utility. In this work, we propose a novel and efficient machine unlearning method for pre-trained models. We term the method Residual Feature Alignment Unlearning. Specifically, we leverage LoRA (Low-Rank Adaptation) to decompose the model's intermediate features into pre-trained features and residual features. By adjusting the residual features, we align the unlearned model with the pre-trained model at the intermediate feature level to achieve both unlearning and remaining targets. The method aims to learn zero residuals on the retained set and shifted residuals on the unlearning set. Extensive experiments on numerous datasets validate the effectiveness of our approach.
\end{abstract}


\section{Introduction}
\IEEEPARstart{M}{achine} unlearning has emerged as a significant research direction in the field of machine learning in recent years, receiving extensive attention for its ability to protect user privacy and remove harmful data. Machine unlearning provides an effective solution to ensure user data can be securely removed from models when requested or no longer needed. It has also been extended to remove harmful or outdated data contained in models, enhancing security and usability.

Current research mainly focuses on how to implement unlearning efficiently and effectively. Deletion requests can be frequent in practice \cite{wang2023kga}, including user requests to remove personal data and the removal of harmful information from models. Therefore, quickly and effectively deleting target data without affecting the model's performance on the retained data is a key focus of current machine unlearning research \cite{nguyen2022survey, musurvey2024}.

Current unlearning methods can be categorized into two types: exact unlearning and approximate unlearning. Exact unlearning requires that the distribution of the unlearned model is indistinguishable from that of a retrained model, while approximate unlearning only requires that the two are approximately indistinguishable \cite{musurvey2024}. Exact unlearning typically requires retraining on all or part of the training data, which incurs high computational and time costs. However, since it can achieve theoretically complete unlearning, it has received significant attention in the early stages of research. A typical method is SISA \cite{bourtoule2021machine}, which stores a checkpoint for each subset during training and only retrains the corresponding sub-model upon receiving a deletion request. However, it requires storing a large number of intermediate results during pre-training, which can consume significant storage space \cite{musurvey2024}. Additionally, it may be ineffective for pre-trained models, which do not partition the data or store intermediate results during pre-training.

To mitigate the above issues, recent work on pre-trained models has shifted towards approximate unlearning. Approximate unlearning does not aim for complete indistinguishability between the unlearned and retrained models but rather for approximate indistinguishability \cite{nguyen2022survey}. It eliminates the influence of the unlearning data by modifying the model’s parameters. Compared to retraining, approximate unlearning only fine-tunes the pre-trained model, reducing computational costs and training time. A typical approach is to compute and remove the contribution of the unlearning data to the model \cite{lin2023erm,guo2019certified}.

\IEEEpubidadjcol

However, existing approximate unlearning methods for pre-trained models still face the following challenges.

\begin{enumerate}
    \item Intermediate feature shift. Current unlearning methods primarily focus on handling the final output, without adequately considering the feature representations in the intermediate layers of the model. In fact, the intermediate features of a pre-trained model are abstract representations learned from a large-scale training set. Removing a small portion of the data may lead to significant shifts in these features, which can impact the model's performance on the retained data.

    \item Smooth Unlearning Transition. The parameters of an unlearned model should ideally remain close to those of the original pre-trained model. However, many fine-tuning-based unlearning methods can cause abrupt and substantial parameter shifts at the start of the process, risking performance degradation on retained data. This is particularly challenging in the early stages, where the model must balance unlearning and retaining. 

    \item Unlearning efficiency. Although approximate unlearning can be achieved through fine-tuning, the computational and memory costs of fine-tuning remain high for large models, making fast and efficient unlearning a significant challenge in practical applications.
\end{enumerate}

To address these challenges, we propose a fast and efficient unlearning method based on residual feature alignment using Low-Rank Adaptation (LoRA) \cite{hu2021lora}. LoRA is a fine-tuning method that introduces low-rank matrices into the model, allowing for efficient updates with low memory consumption and computational overhead. We address the above challenges by introducing LoRA modules in certain intermediate layers of the model: 1) To handle intermediate feature shift, we divide the intermediate features of the model into pre-trained features and residual features. We freeze the pre-trained features and achieve unlearning by adjusting the residual features to align with different objectives on the retained and unlearning sets. 2) The residual features are initialized to zero (or near-zero), ensuring a smoother update process during the initialization of unlearning, preventing the model from deviating too quickly from the pre-trained model in the early stages of unlearning, thus maintaining proximity to the pre-trained model and reducing performance loss. 3) To improve unlearning efficiency, we introduce LoRA modules only in the necessary intermediate layers. By adding low-rank matrices, we significantly reduce memory consumption and computational overhead, enabling even large-scale models to unlearn quickly and efficiently.

Our main contributions are summarized as follows:
\begin{enumerate}
    \item We propose a LoRA-based residual feature alignment method for machine unlearning, which enables efficient unlearning while maintaining consistency between the intermediate features of the unlearned model and the pre-trained model.
    \item We formulate a dual-objective loss function that explicitly aligns residual features for both the retained and unlearning sets, ensuring precise control over the unlearning process. This formulation can be conveniently implemented within a teacher-student framework.
    \item We conducted extensive experiments on various datasets and models. The broad experimental results demonstrate the effectiveness and efficiency of our method. 
\end{enumerate}

\section{Related Work}

\subsection{Machine Unlearning}
Unlearning refers to removing the influence of a specific subset of the training data from a trained model, making the model behave as if it had never seen that subset. The concept of machine unlearning was introduced in \cite{cao2015towards}, where the training dataset is transformed into a summation form, and these transformations are used for training. When unlearning a sample, only the affected sums need to be updated, allowing for fast updates. However, this approach has limited applicability and is restricted to structured problems \cite{sekhari2021remember}. The classic SISA method \cite{bourtoule2021machine} partitions the training set into non-overlapping shards and creates sub-models for each. Inspired by the definition of differential privacy \cite{dwork2006differential}, \cite{ginart2019making} introduced the probabilistic notion of unlearning, aiming for the distribution of the unlearned model to be similar to that of the retrained model. Based on the comparison of these two distributions \cite{thudi2022unrolling,guo2019certified,izzo2021approximate,neel2021descent}, the objectives of unlearning can be classified into exact unlearning and approximate unlearning \cite{nguyen2022survey}. These distributions can be represented in both weight space and output space \cite{nguyen2022survey}.

Exact unlearning ensures that the distributions of the unlearned model and the retrained model are indistinguishable, which is often challenging to achieve \cite{musurvey2024}. SISA \cite{bourtoule2021machine} is considered exact unlearning in some works \cite{lin2023erm}, but the new model obtained through shard retraining is clearly not similar to the traditional non-sharded retrained model. Approximate unlearning, a compromise of exact unlearning, restricts the divergence between the two distributions within a tolerable range, typically based on the definition of differential privacy \cite{dwork2006differential}. Many related works are based on the concept of approximate unlearning \cite{graves2021amnesiac,golatkar2020eternal,guo2019certified,sekhari2021remember,warnecke2021machine,thudi2022unrolling}. Representative methods include gradient ascent \cite{graves2021amnesiac}, fisher unlearning \cite{golatkar2020eternal}, and influence unlearning \cite{guo2019certified}. Furthermore, \cite{thudi2022necessity} points out that different samples can generate the same network, suggesting that the definition of approximate unlearning is flawed and fails to provide proof of unlearning.

There are two main unlearning scenarios based on the nature of the unlearning set: class unlearning and sample unlearning. Class unlearning refers to the removal of all samples from a specific category, which is typically used to delete an entire category of data from the model \cite{tarun2023fast,graves2021amnesiac,chundawat2023zero,yoon2022few}. Sample unlearning involves unlearning random samples from various categories, usually used to remove specific data points from the model \cite{cha2024learning,golatkar2021mixed,mehta2022deep,kim2022efficient}.


\subsection{LoRA}
LoRA (Low-Rank Adaptation) \cite{hu2021lora} is an efficient fine-tuning method that reduces the number of parameters and computational resources required to update large pre-trained models by introducing low-rank matrices. Specifically, high-dimensional matrices are represented as the product of two low-rank matrices to simulate parameter updates. During fine-tuning, the parameters of the pre-trained model are frozen, and only the parameters of the low-rank matrices are adjusted, allowing the model to adapt to new tasks while preserving the core capabilities of the pre-trained model. Additionally, during inference, the pre-trained weights and the low-rank matrices can be merged, avoiding additional inference latency \cite{hu2021lora}. We extend the concept of LoRA by viewing the output of the low-rank matrices as the residual features between the unlearned model and the original model. By adjusting the residual features, we align the unlearned model with the feature differences between the retained and unlearning sets, thereby achieving unlearning.

\section{Preliminary}

\subsubsection{Notation}
Let \( \mathcal{D} = \{(z_i, y_i)\}_{i=1}^N \) denote a training set containing \( N \) samples, where each \( z_i \) represents the input sample features and \( y_i \) is the corresponding label. Let \( f \) denote a deep neural network with \( m \) layers, which is a composite function of the set of functions \( \mathcal{F} = \{f_i\}_{i=1}^m \), i.e., \( f = f_m(f_{m-1}(...f_2(f_1(\cdot)))) \). The parameters corresponding to the functions in \( \mathcal{F} \) are denoted as \( w = \{w_1, w_2, ..., w_m\} \). For the sake of discussion, we do not consider non-trainable layers and their parameters. In a deep neural network, the original input features are transformed and passed through layers. We use \( x_k \) to denote the output features of the \( k \)-th layer, and specifically, we use \( x_0 \) to denote the original input features of the sample.

\subsubsection{Machine Unlearning}
Machine unlearning is defined as the ability to remove the influence of specified training data from a model while preserving its original performance. Let \( \mathcal{D}_f = \{(z_i, y_i)\}_{i=1}^{N_f} \) and \( \mathcal{D}_r = \{(z_i, y_i)\}_{i=1}^{N_r} \) denote complementary subsets of the training set \( \mathcal{D} \), containing \( N_f \) and \( N_r \) samples respectively, such that \( \mathcal{D}_f \cup \mathcal{D}_r = \mathcal{D} \) and \( \mathcal{D}_f \cap \mathcal{D}_r = \varnothing \), where \( \varnothing \) represents the empty set. Here, \( \mathcal{D}_f \) represents the unlearning data, and \( \mathcal{D}_r \) represents the retained data. Typically, \( N_r \gg N_f \). Suppose \( f \) is a model trained on \( \mathcal{D} \); machine unlearning aims to remove the influence of the data samples in \( \mathcal{D}_f \) from \( f \) while preserving its utility and performance on \( \mathcal{D}_r \). Based on the samples in \( \mathcal{D}_f \), there are two main unlearning scenarios: class unlearning and sample unlearning. Class unlearning refers to the case where the samples in \( \mathcal{D}_f \) comprise all samples of a specific class, which is commonly used to delete data for an entire class from the model. Sample unlearning refers to the case where the samples in \( \mathcal{D}_f \) are random samples of unspecified classes, typically used to remove certain specific data points from the model.

\subsubsection{Low-Rank Adaptation}
LoRA is an efficient fine-tuning method that reduces the complexity and computational cost of adjusting model parameters by introducing low-rank matrices. Let \( W \in \mathbb{R}^{d \times v} \) represent the pre-trained weight matrix, and \( x_k \) represent the hidden features of the \( k \)-th layer. In LoRA, the update of \( W \) is decomposed into two low-rank matrices, namely \( \Delta W = BA \), where \( B \in \mathbb{R} ^{d \times u} \) and \( A \in \mathbb{R}^{u \times v} \), with the rank \( u \ll \min(d, v) \). LoRA initializes \( B \) and \( A \) such that \( \Delta W = BA \) is zero at the beginning of training. During fine-tuning, the pre-trained weight \( W \) is frozen, and only \( B \) and \( A \) are updated. As shown in Fig.~\ref{fig:lora}, for \( x_k = W x_{k-1} \), the modified forward pass is:
\[ x_k = x_k' + \Delta x_k' = W x_{k-1} + BA x_{k-1} \]
LoRA simulates the process of full-parameter fine-tuning by updating additional matrices. Compared to full-parameter fine-tuning, LoRA requires training only the smaller low-rank matrices, significantly reducing the number of trainable parameters and thereby substantially lowering the computation and memory overhead during fine-tuning. Moreover, during inference, the matrices \( W \) and \( BA \) can be merged without incurring extra inference latency.

\section{Residual Feature Alignment Unlearning}

\subsection{Method Overview}

We propose an unlearning method based on residual feature alignment, viewing unlearning as a shift of the original model’s features following a specific pattern. This shift is achieved through residual features, ensuring the consistency of the model's feature extraction ability before and after unlearning.

\begin{figure}[!t]
\centering
\includegraphics[width=0.96\columnwidth]{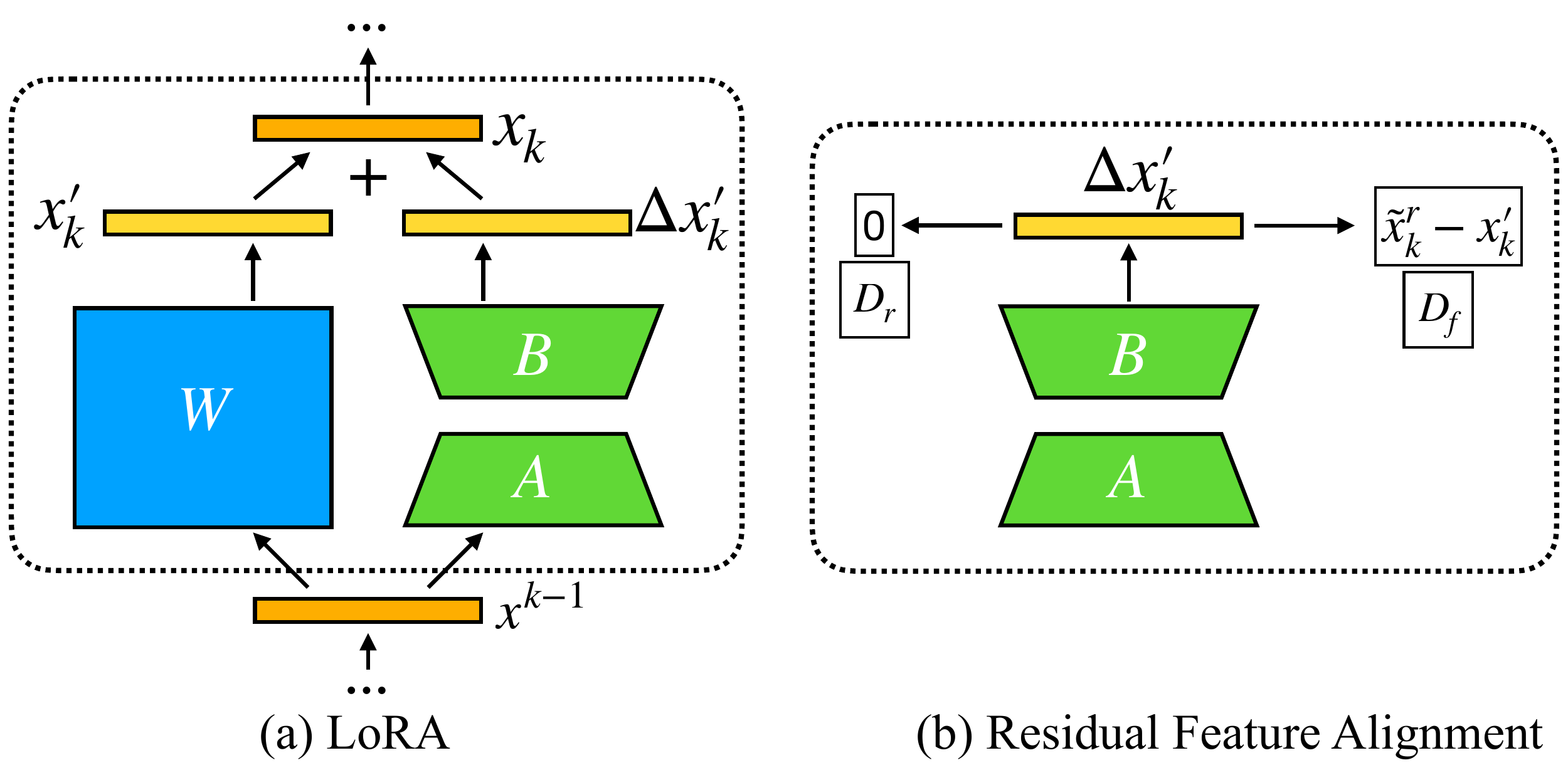}
\caption{(a) and (b) illustrate the residual feature alignment based on LoRA. (a) depicts the training process of LoRA, where the pre-trained weights \( W \) are frozen, and only \( A \) and \( B \) are trained. Unlike the original LoRA, we take the output \( \Delta x_{k}' \) of the \( BA \) branch as our target. In (b), different targets are set for \( \Delta x_{k}' \) on \( \mathcal{D}_r \) and \( \mathcal{D}_f \), respectively, to achieve retention and unlearning of the model in the intermediate features.}
\label{fig:lora}
\end{figure}

First, we provide a detailed explanation of the residual feature alignment mechanism. We decompose the model's features into pre-trained features and residual features, obtained from the pre-trained branch and the bypass branch, respectively, as shown in Fig.~\ref{fig:lora}. The bypass branch is implemented using LoRA. By adjusting only the residual features, we dynamically tune the features at each layer. We discuss different adjustment strategies for the retained and unlearning sets, analyze the essence of residual feature alignment, and explain why LoRA is used to implement residual feature alignment.

Next, we analyze in detail how residual feature alignment can be used to achieve unlearning. We add a bypass branch composed of low-rank matrices at specific intermediate layers, with the optimization objective set on the output of the bypass branch. Let the function of the original model’s \( k \)-th layer be \( f_k \), the function of the bypass branch be \( f_k' \), and the input be \( x_{k-1} \), then the new function at the \( k \)-th layer is:

\[
x_k = f_k(x_{k-1}) + f_k'(x_{k-1})
\]

\begin{figure}[!t]
\centering
\includegraphics[width=0.98\columnwidth]{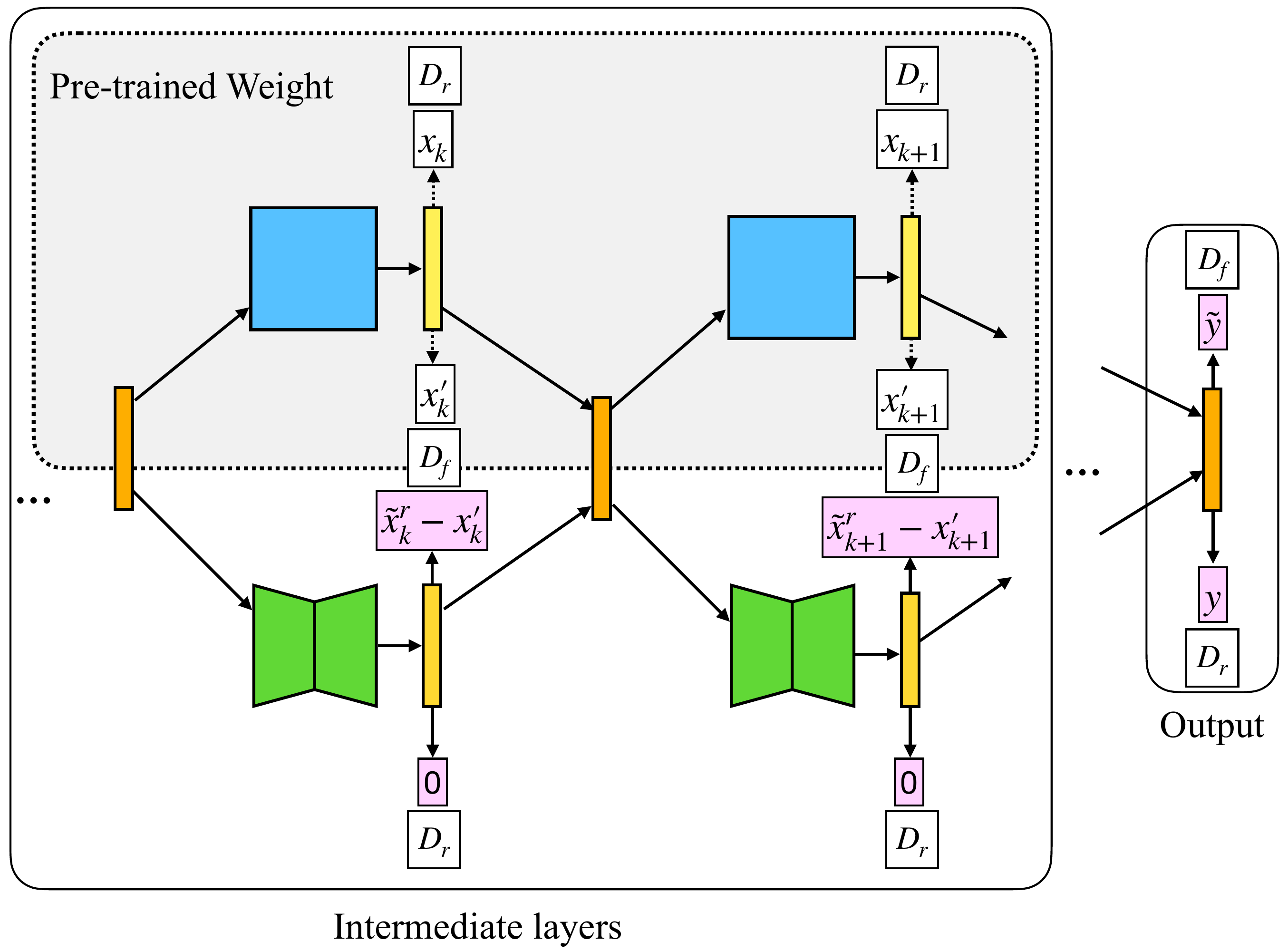}
\caption{The unlearning process of residual feature alignment. During training, we freeze the pre-trained weights and train only the incremental network added to the intermediate layers. We spread the optimization objective across the intermediate layers. For the retained set \( \mathcal{D}_r \), we aim for the output of each layer's incremental weights to be zero, thereby ensuring that the output of the pre-trained weights on \( \mathcal{D}_r \) remains unaffected. For the unlearning set \( \mathcal{D}_f \), we aim for the output of each layer’s incremental weights to match the residual between the pre-trained weights and the average feature on \( \mathcal{D}_r \), achieving the purpose of unlearning. At the output layer, we adopt a similar averaging strategy for \( \mathcal{D}_f \).}
\label{fig:lora_residual}
\end{figure}

During the model unlearning process, the loss function corresponding to the \( k \)-th layer is:

\[
L_k = \|f_k'(x_{k-1}) - t^k\|^2
\]

where \( t^k \) denotes a specific target value, which is chosen differently for the retained set and the unlearning set. The entire unlearning process is illustrated in Fig.~\ref{fig:lora_residual}.

As shown in Fig.~\ref{fig:lora_residual}, we essentially optimize the incremental network of the original model, using this incremental network to achieve the model's unlearning process. We also extend this architecture to a teacher-student framework, as depicted in Fig.~\ref{fig:lora_teacher}, to facilitate code implementation.

\subsection{Residual Feature Alignment}\label{sec:res_feat_align}

Deep learning models transform input data in the original space through multiple layers to obtain intermediate features, which are then passed to subsequent layers. These intermediate features often contain abstract information about the original data \cite{adriana2015fitnets, kingma2013auto, rombach2022high}. The goal of unlearning is to forget information in the original space, and a common approach is to fine-tune the logits of the unlearning samples \(\mathcal{D}_f\) at the output layer \cite{kurmanji2024towards, chundawat2023can}. However, this is often challenging and resource-intensive. Since the original space contains a large amount of redundant information, directly operating in this space can lead to unconstrained adjustments in many intermediate layers, making the intermediate features in the retained set inconsistent after unlearning, which may affect the normal utility of the pre-trained model.

Unlike directly fine-tuning in the original space, we modify specific intermediate layers in the feature space, aligning the intermediate features according to a specific pattern to achieve unlearning while preserving the original model's features. The alignment is achieved through the residuals relative to the original weights, for which we introduce LoRA. Specifically, during the unlearning process, we divide the intermediate features into two parts: pre-trained features and residual features. The pre-trained features correspond to the model's original feature extraction capability, obtained through frozen pre-trained weights \(W\); the residual features correspond to the feature extraction capability learned during unlearning, obtained through the trainable LoRA weights \(BA\).

As shown in Fig.~\ref{fig:lora}(a), the features of a certain layer, \(x_{k-1}\), are passed through both the pre-trained module and the LoRA module to obtain the next layer's features: \(x_k = x_k' + \Delta x_k'\), where \(x_k' = W x_{k-1}\) represents the pre-trained features, and \(\Delta x_k' = BA x_{k-1}\) represents the residual features. We achieve unlearning by aligning \(\Delta x_k'\) to different target values for the retained and unlearning sets, as illustrated in Fig.~\ref{fig:lora}(b):

\[
\begin{cases}
\Delta x_k' \to 0, & \text{for } \mathcal{D}_r \\
\Delta x_k' \to \textit{specific target}, & \text{for } \mathcal{D}_f
\end{cases}
\]

For the retained data, we aim for \(x_k = x_k' + \Delta x_k' \approx x_k'\), i.e., aligning \(\Delta x_k'\) to $0$ to avoid impacting the model's feature extraction capability on the retained data. For the unlearning data, we aim for \(x_k = x_k' + \Delta x_k' \neq x_k'\), and for this, we design the target value of \(\Delta x_k'\) to be a specific pattern, changing the feature distribution on the unlearning data to achieve unlearning. This specific pattern can be arbitrary, but it is usually chosen to be a distribution related to the pre-trained features \(x_k'\).

\subsubsection{Alignment Pattern on Retained Data}

For the retained data, we only need to maintain the model's original feature distribution, which means we need to set the target value of \( \Delta x_k' \) to $ 0 $. This is not an easy target to achieve with completely random initialization. Fortunately, the initialization method of LoRA ensures that this target is achieved at the beginning of training. However, it is important to note that at this stage, the bypass branch has not yet learned any information about the retained data. It simply outputs $ 0 $ because the matrix \( B \) is initialized to $ 0 $. In subsequent training, the bypass branch needs to truly learn the information from the retained data and map it to $ 0 $. Theoretically, optimization on the unlearning data could affect the performance on the retained data, causing \( \Delta x_k' \) to gradually deviate from $ 0 $ (though this deviation is typically small since the retained set is usually larger). Therefore, we need to design a special distribution shift for the model features on the unlearning data.

\subsubsection{Alignment Pattern on Unlearning Data}

\begin{figure}[t]
\centering
\includegraphics[width=0.96\columnwidth]{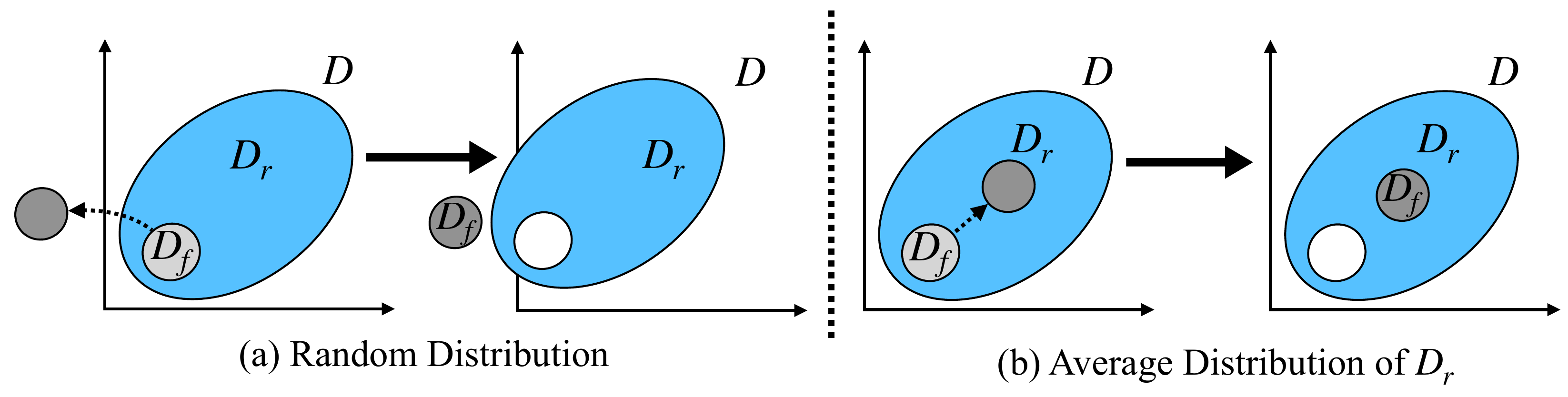}
\caption{(a) and (b) illustrate the impact on features in \( \mathcal{D}_r \) when transferring features from \( \mathcal{D}_f \) to different distributions. In (a), transferring the intermediate features from \( \mathcal{D}_f \) to a random distribution may cause a shift in the features of \( \mathcal{D}_r \) as well. In (b), transferring the intermediate features from \( \mathcal{D}_f \) to the average distribution of \( \mathcal{D}_r \) ensures that the target feature distributions on \( \mathcal{D}_f \) and \( \mathcal{D}_r \) are similar, thereby reducing the impact on the feature distribution in \( \mathcal{D}_r \).}
\label{align-pattern}
\end{figure}

For the unlearning data, we need to change the feature distribution of the model. A naive approach would be to shift the feature distribution of the unlearning data \( \mathcal{D}_f \) to another random distribution. However, this could alter the feature distribution on the retained set \( \mathcal{D}_r \), even though we set the goal \(\Delta x_k' \rightarrow 0\) for \( \mathcal{D}_r \). As shown in Fig.~\ref{align-pattern}(a), the features of \( \mathcal{D}_f\) are randomly shifted to another distribution, completely altering the features on \(\mathcal{D}_f\). But during training, to fit the changes on \(\mathcal{D}_f\), the features on \(\mathcal{D}_r\) also change, which is undesirable.

To mitigate this, we shift the features of \(\mathcal{D}_f\) towards the average distribution of \(\mathcal{D}_r\), as shown in Fig.~\ref{align-pattern}(b). In this way, the target feature distributions on \(\mathcal{D}_f\) and \(\mathcal{D}_r\) become more aligned, reducing the impact on \(\mathcal{D}_r\)'s features.

For convenience, let \(x_f\) and \(\Delta x_f\) represent the pre-trained features and residual features of a certain layer on \(\mathcal{D}_f\). The target value of the residual features on \(\mathcal{D}_f\) is:

\[ \Delta x_f \rightarrow (\frac{1}{N_r} \sum_{x \in \mathcal{D}_r} x_r) - x_f \]

where \(x_r\) represents the pre-trained features of the corresponding layer in \(\mathcal{D}_r\), and \(N_r\) is the number of samples in \(\mathcal{D}_r\). The residual features are added to the pre-trained features during model inference, so that the summed features on \(\mathcal{D}_f\) are close to the average features on \(\mathcal{D}_r\), thereby achieving unlearning while preserving the model's utility.

\subsubsection{Analysis}

As shown in Fig.~\ref{fig:lora_residual}, all the LoRA modules together form a new network, which is essentially the incremental network of the original model. The target values of this incremental network at each intermediate layer are only two: $ 0 $ and the residual of the average features, corresponding to the retained set and the unlearning set, respectively. The alignment refers to the incremental network mapping the differences between the two to a $ 0 $ distribution and an average feature residual distribution. If the incremental network were to learn the original data's features from scratch, it would be a difficult task. However, in unlearning, the incremental network only needs to fit the differences between these two datasets, making the training process much simpler.

\subsubsection{Why Use LoRA for Residual Features}

We aim to implement unlearning at the feature level, not just at the output level. This requires modifying the features of the unlearning dataset while preserving those of the retained dataset. While various parameter-efficient fine-tuning (PEFT) methods exist, we identify LoRA as particularly well-suited for this task for the following reasons:
1) Learning Residual Updates. LoRA's core principle is to learn a low-rank update, $\Delta W = BA$, which naturally captures the ``residual'' change we aim to induce. While other methods like PiSSA \cite{meng2024pissa} or DoRA \cite{liu2024dora} also introduce adaptable parameters, our ablation study (see Section \ref{sec:ablation_peft}) demonstrates that LoRA achieves the optimal balance between unlearning efficacy and feature space stability.
2) Stable Zero Initialization. LoRA's initialization, where matrix $B$ is set to zero, ensures that the residual feature is zero at the start of the unlearning process ($\Delta x_k' = 0$). This provides a smooth and stable starting point, preventing drastic shifts in the model's behavior on the retain set.
3) Computational Efficiency and Generality. LoRA is highly computationally efficient and architecture-agnostic. Unlike methods such as LLaMA-Adapter \cite{gao2023llama}, which relies on appending learnable prompts to token sequences and is inherently designed for attention-based architectures, LoRA can be universally applied to CNNs, Transformers, and LLMs. Furthermore, LLaMA-Adapter may introduce additional inference latency. In contrast, LoRA allows the matrix $BA$ to be merged with the pre-trained weights $W$ ($W' = W + BA$) after training. The benefit of this is that the model doesn't require extra computational layers or parallel paths during inference, ensuring zero additional inference latency.

\subsection{Residual Feature Alignment for Unlearning}

We aim to impose constraints on the model's hidden features during unlearning, and these constraints are applied layer by layer through residual feature alignment. For a given intermediate layer, we seek to learn the zero residual on \(\mathcal{D}_r\) and the shifted residual on \(\mathcal{D}_f\). Since we freeze the pre-trained weights \(W\) during training, our training objective focuses only on the residual features. Specifically, suppose the network has \(m\) layers (for simplicity, we do not consider non-trainable layers such as Dropout, Normalization, etc.). For the \(k\)-th layer, the residual feature is:

\[ \Delta x_k' = B_k A_k x_{k-1} \]

where \(B_k\) and \(A_k\) are the two low-rank matrices of the LoRA module,  with the subscript \(k\) distinguishing the parameters of different layers. Initially, \(\Delta x_k'=0\) because the \(B_k\) matrix is initialized to $ 0 $. Therefore, at the start of training, although the LoRA modules are added to different layers, the features of each layer remain identical to those of the original model.

This initialization method is crucial for the training on \(\mathcal{D}_r\), as it makes it easier to achieve our objective on \( \mathcal{D}_r \), which is to maintain consistency between the features of the unlearned model and the original model on \( \mathcal{D}_r \). To continue preserving this objective during subsequent training on \(\mathcal{D}_r\), our optimization objective is:

\begin{align}
\min \frac{1}{N_r} \sum_{x \in \mathcal{D}_r} \sum_{k=1}^m ||\Delta x_k'||
\label{eq:objective_dr}
\end{align}

For \(\mathcal{D}_f\), we need to shift the feature distribution of the original model to achieve unlearning. In the analysis above, we chose to shift the original features on \(\mathcal{D}_f\) towards the average features on \(\mathcal{D}_r\). Specifically, we use the average features of all \(\mathcal{D}_r\) samples in a batch at the \(k\)-th layer as the target value for the corresponding layer features in \(\mathcal{D}_f\). Let the average features of \(\mathcal{D}_r\) at \(k\)-th layer be \( \tilde{x}_k^r = \frac{1}{N_r} \sum_{x \in \mathcal{D}_r} x_k \). Then, the optimization objective for \(\mathcal{D}_f\), applied to the residual, is:

\begin{align}
\min \frac{1}{N_f} \sum_{x \in \mathcal{D}_f} \sum_{k=1}^m ||\Delta x_k' - (\tilde{x}_k^r - x'_k)||
\label{eq:objective_df}
\end{align}

The objectives for both \(\mathcal{D}_r\) and \(\mathcal{D}_f\) ensure feature-level consistency. 
To ensure consistency at the model's output layer, which can be viewed as the final feature representation, we extend our alignment strategy to the model's predictions. For the retained set $\mathcal{D}_r$, we use the standard task loss with the ground-truth labels. For the unlearning set $\mathcal{D}_f$, simply using the original labels would contradict the unlearning objective. Instead of assigning random labels, which can introduce noisy signals and destabilize the model's learned distributions, we propose aligning the output for $\mathcal{D}_f$ samples with the model's own average prediction distribution on the retained data $\mathcal{D}_r$.

Specifically, for a batch of data, we compute the average softmax probability distribution over all samples in the retained set of the batch. This average distribution, denoted as $\tilde{y}$, serves as a soft target for the samples in the unlearning set. 
The task loss for the unlearning set then becomes a distribution matching objective, encouraging the model's predictions on the unlearning set to align with its average predictions on the retained set. As shown in Fig.~\ref{align-pattern}, this method smoothly guides the model to unlearn by following the output distribution of the retained data. This makes the outputs on the unlearning set similar to those on the retained set, which prevents a dramatic change in the model's output space for the retained set. The corresponding loss term is:

\begin{align}
\min \frac{1}{N_r} \sum_{x \in \mathcal{D}_r} l(f(x), y) + \frac{1}{N_f} \sum_{x \in \mathcal{D}_f} l(f(x), \tilde{y})
\label{eq:objective_task}
\end{align}

where $l(\cdot, \cdot)$ is the model's task loss function, and $\tilde{y}$ is the model's average prediction on the retained set $\mathcal{D}_r$, i.e., $\tilde{y}=\frac{1}{N_r} \sum_{x \in \mathcal{D}_r} f(x)$.

In the optimization objectives above, terms \ref{eq:objective_dr} and \ref{eq:objective_df} represent intermediate layer losses, while term \ref{eq:objective_task} represents the task loss:

\begin{align}
L_{inter} = & \frac{\alpha}{N_r} \sum_{x \in \mathcal{D}_r} \sum_{k=1}^m ||\Delta x_k'|| \notag \\
    & + \frac{\beta}{N_f} \sum_{x \in \mathcal{D}_f} \sum_{k=1}^m ||\Delta x_k' - (\tilde{x}_k^r - x'_k)||
\end{align}

\begin{align}
L_{task} = & \frac{\lambda}{N_r} \sum_{x \in \mathcal{D}_r} l(f(x), y) + \frac{\mu}{N_f} \sum_{x \in \mathcal{D}_f} l(f(x), \tilde{y})
\end{align}

Our final optimization objective is defined as:

\begin{align}
L = & \frac{\gamma}{m} L_{inter} + (1 - \gamma) L_{task}
\end{align}

where \( \alpha \), \( \beta \), \( \lambda \), \( \mu \) and \( \gamma \) are hyperparameters. We summarize the residual feature alignment for unlearning in Algorithm \ref{alg:algorithm1}.

\begin{algorithm}
\caption{Residual Feature Alignment Unlearning}
\label{alg:algorithm1}
\begin{algorithmic}[1]
\STATE \textbf{Input:} The full training set $\mathcal{D}$, unlearning set $\mathcal{D}_f$, retained set $\mathcal{D}_r$, original model weights $W^o$
\STATE \textbf{Output:} Unlearning model weights $W^u$
\STATE Initialize $W^u \gets W^o$
\STATE Insert LoRA into the intermediate layers of $W^u$
\FOR{epoch $e = 1$ to $E$}
    \FOR{batch $b \in \mathcal{D}$}
        \STATE $\Delta x_k' \gets$ Output of the $k$-th LoRA module
        \STATE $x_k \gets$ Output of the $k$-th pre-trained module
        \STATE $
        \begin{aligned}[t]
        \tilde{x}_k^r \gets \frac{1}{|b_r|} \sum_{x \in b_r}x_k, \tilde{y} \gets \frac{1}{|b_r|} \sum_{x \in b_r} f(x)
        \end{aligned}
        $
        \STATE $
        \begin{aligned}[t]
        L_{\text{inter}} \gets{}& \frac{\alpha}{|b_r|}\sum_{x\in b_r} \sum_{k=1}^{m} \| \Delta x_k' \| \\
        & + \frac{\beta}{|b_f|} \sum_{x \in b_f}\sum_{k=1}^{m} \| \Delta x_k' - ( \tilde{x}_k^r - x'_k ) \|
        \end{aligned}
        $
        \STATE $
        \begin{aligned}[t]
        L_{\text{task}} \gets{}& \frac{\lambda}{|b_r|}\sum_{x\in b_r} l(f(x), y) + \frac{\mu}{|b_f|}\sum_{x \in b_f} l(f(x), \tilde{y})
        \end{aligned}
        $
        \STATE $L \gets \gamma/m \cdot L_{\text{inter}} + (1 - \gamma) \cdot L_{\text{task}}$
        \STATE $W^u \gets W^u - \epsilon \nabla L$
    \ENDFOR
\ENDFOR
\STATE Merge LoRA weights into the pre-trained weights
\STATE \textbf{Return:} $W^u$
\end{algorithmic}
\end{algorithm}

\subsubsection{Training Architecture}

The above objective function can become complex to implement in code, as it requires separately computing the residual features at each layer. This means that in the forward pass of modules containing LoRA branches, both the pre-trained features and residual features need to be returned separately, rather than simply adding them together and passing the sum to the next layer. To ensure the proper functioning of the subsequent layers, some special programming techniques are required, such as implementing the process of \( x_k = x_k' + \Delta x_k' \) in hook functions. This increases the complexity of the code implementation.
In the specific code implementation, we can transform the loss functions for easier coding. The loss functions that need to be transformed are given in Equations \ref{eq:objective_dr} and \ref{eq:objective_df}. Let \( x_k^s = \Delta x'_{k} + x'_k \). Then, for Equation \ref{eq:objective_dr}:
\begin{align}
&\frac{1}{N_r} \sum_{x \in \mathcal{D}_r} \sum_{k=1}^m \| \Delta x'_{k} \| \notag \\
&= \frac{1}{N_r} \sum_{x \in \mathcal{D}_r} \sum_{k=1}^m \| (\Delta x'_{k} + x'_k) - x'_k \|\notag \\
&= \frac{1}{N_r} \sum_{x \in \mathcal{D}_r} \sum_{k=1}^m \| x_k^s - x'_k \|
\end{align}
For Equation \ref{eq:objective_df}:

\begin{align}
&\frac{1}{N_f} \sum_{x \in \mathcal{D}_f} \sum_{k=1}^m \ \| \Delta x'_{k} - ( \tilde{x}_{k}^r - x'_{k} ) \| \notag \\
&= \frac{1}{N_f} \sum_{x \in \mathcal{D}_f} \sum_{k=1}^m \| (\Delta x'_{k} + x'_{k}) - \tilde{x}_{k}^r \| \notag \\
&= \frac{1}{N_f} \sum_{x \in \mathcal{D}_f} \sum_{k=1}^m \| x_k^s - \tilde{x}_{k}^r \|
\end{align}

\begin{figure}[t]
\centering
\includegraphics[width=0.96\columnwidth]{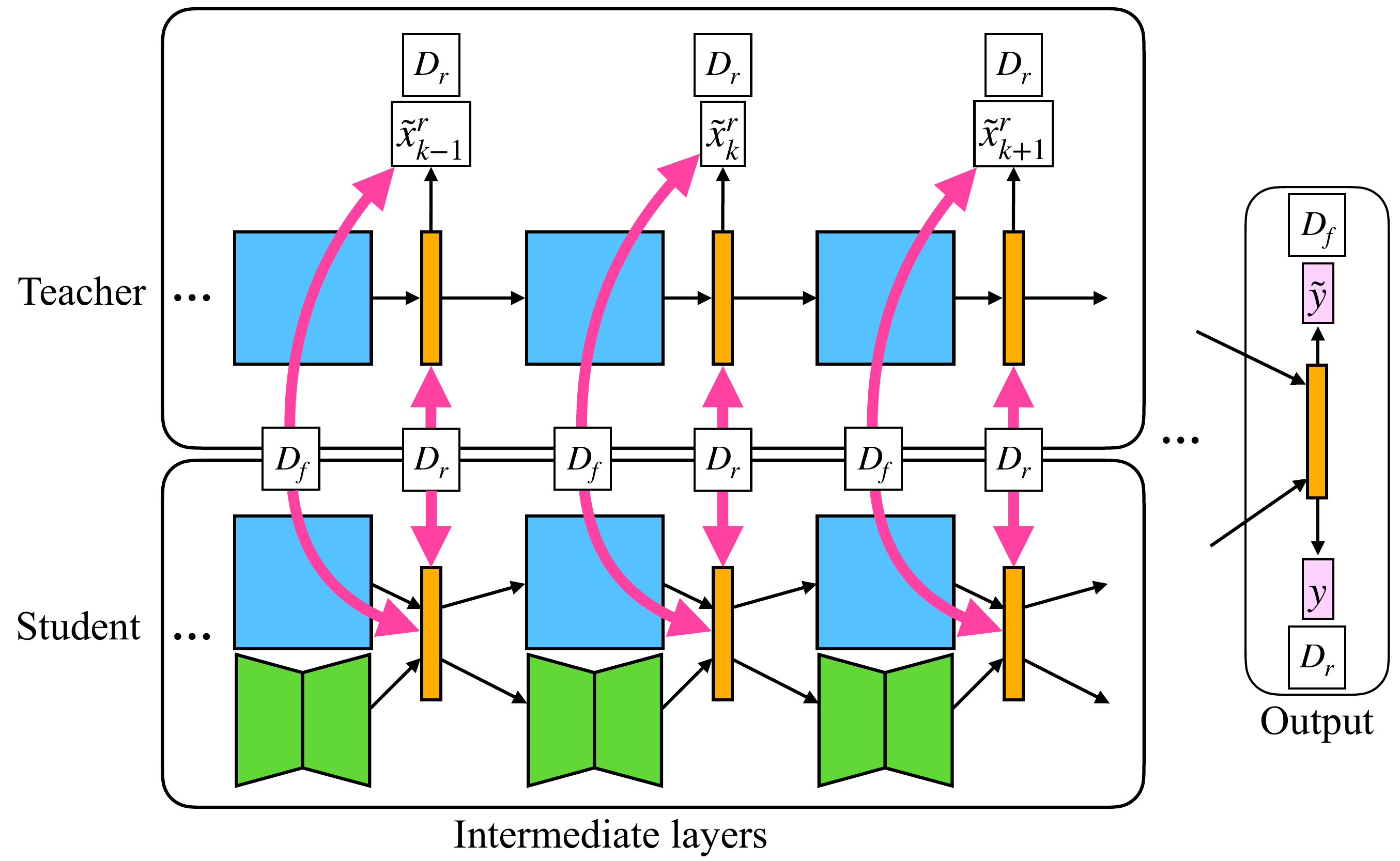}
\caption{ To simplify code implementation, a teacher-student network architecture can be used for training. The original model serves as the teacher model, while the unlearning model with LoRA serves as the student model. On \(\mathcal{D}_r\), the student model aligns its intermediate features with the corresponding features of the teacher model. On \(\mathcal{D}_f\), the student model aligns with the average intermediate features of the teacher model obtained on \(\mathcal{D}_r\).}
\label{fig:lora_teacher}
\end{figure}

Therefore, we can follow the feature-based distillation \cite{gou2021knowledge} training architecture, as illustrated in Fig.~\ref{fig:lora_teacher}.

Specifically, we align the intermediate layers of the unlearning model with those of the original model. On the retained set \(\mathcal{D}_r\), the intermediate features of the unlearned model (which consist of the sum of pre-trained and residual features) are directly aligned with the intermediate features of the original model. On the unlearning set \(\mathcal{D}_f\), the intermediate features of the unlearned model are aligned with the average features of the original model. During unlearning, since the pre-trained weights are frozen and not updated, the pre-trained features remain fixed. Therefore, the gap between the intermediate features of the unlearned model and the original model can only be filled by the residual features.

During training, the forward pass of modules with LoRA branches does not need to return pre-trained and residual features separately; instead, the two are directly summed, reducing the complexity of code implementation. The trade-off is that the unlearning process requires the additional output of the teacher network, which increases computational and memory overhead.

\section{Experiments}

In this section, we conduct extensive experiments to validate our proposed method. Our primary evaluation criterion is to measure how closely the unlearned model, generated by our method, approximates the behavior of a model retrained from scratch on the retained data ($\mathcal{D}_r$). The `Retrain' model serves as the gold standard for successful unlearning \cite{musurvey2024, nguyen2022survey}. Therefore, across all utility, privacy, and efficiency metrics, our goal is to demonstrate that our method's performance is consistently closer to the `Retrain' model than other approximate unlearning baselines.

\subsection{Experimental Setup}

\subsubsection{Tasks}
Our unlearning algorithm is designed to be general, not specific to any particular model architecture. Specifically, we conducted experiments on three common tasks: image classification, text classification, and text generation. We followed the approach in \cite{chundawat2023can,cha2024learning} and performed image classification on two architectures: CNN and transformer. We included two NLP tasks: text classification and text generation. Text classification involves assigning labels or categories to text. We performed experiments on the commonly used sentiment analysis task in text classification. Text generation refers to generating responses given some input content, and we used the causal language modeling. Furthermore, we validated the generalizability of our method on large language models.

\subsubsection{Datasets}
We used different datasets for image classification, text classification, and text generation tasks. For image classification tasks, we used the CIFAR-10\cite{krizhevsky2009learning} and Fashion-MNIST \cite{xiao2017fashion} datasets. For text classification tasks, we utilized the IMDB dataset\cite{maas-EtAl:2011:ACL-HLT2011}, a sentiment classification dataset comprising user reviews of movies. 
We selected samples containing certain keywords in user reviews as the target unlearning class, ensuring that samples with the chosen keyword had a similar proportion in both the training and test sets, with the unlearning samples accounting for $5\%$ to $10\%$ of the total samples. For text generation tasks, we used the ELI5-Category dataset, an English dataset containing questions and answers. We selected $10,000$ samples from $10$ categories, ensuring a relative balance in the number of samples per category, and manually divided them into training and test sets. For large language models, we used the WMDP \cite{li2024wmdp} dataset, which is a dataset designed to measure hazardous knowledge in the fields of biosecurity, cybersecurity, and chemical security.

\subsubsection{Models} 

For image classification tasks, we used the ResNet18 \cite{he2016deep} and Vision Transformer (ViT) \cite{dosovitskiy2020image} models. These models represent classic CNN and transformer architectures for image classification tasks, respectively. To expedite training, we initialized the models with weights pre-trained on ImageNet \cite{imagenet15russakovsky} and then fine-tuned them on the CIFAR-10 and Fashion-MNIST datasets to obtain the original models. For text classification, we followed \cite{wang2023kga,mehta2022deep}, using the DistilBERT \cite{Sanh2019DistilBERTAD}, a distilled version of BERT \cite{devlin2018bert}. In text generation tasks, we used the distilled version of GPT-2 \cite{radford2019language}, called DistilGPT2 \cite{Sanh2019DistilBERTAD}. For DistilBERT and DistilGPT2, we fine-tuned on IMDB and ELI5-Category datasets, respectively, to obtain original models for our experiments. Regarding large language models, we utilized Qwen2.5-3B \cite{qwen2.5} and fine-tuned it on the WMDP \cite{li2024wmdp} dataset to obtain the original model.

\subsubsection{Unlearning Scenarios}

We conducted experiments on two common unlearning scenarios: 1) class unlearning \cite{graves2021amnesiac} and 2) sample unlearning \cite{cha2024learning}. In class unlearning, in tasks with explicit labels (e.g., image classification), we randomly selected a class from the training set as the unlearning target. For datasets without specific classes related to the unlearning task, such as IMDB, we randomly selected target keywords from the training set as labels, treating samples containing the keywords as a single class and making it the unlearning target. Notably, since IMDB has only two labels (positive and negative), it isn’t suitable for use as a target class for unlearning. In sample unlearning, we randomly selected a specific number of samples from the training set as unlearning samples. To ensure experimental consistency, we used the same unlearning samples when evaluating the effects of different models on the same task.

In these two unlearning scenarios, the subsets of interest differ. For class unlearning, we typically focus on four subsets: the retained set and the unlearning set in the training set, denoted as \( \mathcal{D}_r \) and \( \mathcal{D}_f \), respectively, and the retained set and the unlearning set in the test set, denoted as \( \mathcal{D}_{rt} \) and \( \mathcal{D}_{ft} \), respectively. This is because the test set may also contain the target class that needs to be forgotten. In sample unlearning, since the test set usually does not include the samples that need to be forgotten, we only need to focus on the retained set \( \mathcal{D}_r \) and the unlearning set \( \mathcal{D}_f \) in the training set, as well as the test set \( \mathcal{D}_t \).

\subsubsection{Evaluation Metrics}

For unlearning methods, multiple metrics are typically required to evaluate the effectiveness of unlearning. Based on \cite{chen2023boundary}, we categorize metrics into utility metrics, privacy metrics, and computational metrics. Our evaluation includes the following: 
1) \textbf{Accuracy}: For image and text classification tasks, the accuracy of the unlearning model on \( \mathcal{D}_f \), \( \mathcal{D}_r \), \( \mathcal{D}_{rt} \), and \( \mathcal{D}_{ft} \) should be similar to that of the retrained model. 
2) \textbf{Perplexity}: For text generation tasks using causal language models, perplexity reflects the model’s predictive ability on a given dataset. The unlearning model's perplexity should be comparable to that of the retrained model.
3) \textbf{Activation Distance}: Activation distance measures the distance between the final activation values of the unlearning model and the retrained model. A smaller activation distance indicates a stronger unlearning ability. 
4) \textbf{Feature Distance}: Feature distance measures the distance between the intermediate layers of the unlearning and original (or retrained) models. Ideally, the unlearning model should exhibit a feature distance on \( \mathcal{D}_f \) and \( \mathcal{D}_r \) that is comparable to that of the retrained model.
5) \textbf{Membership Inference Attack (MIA)}: We use MIA to investigate whether information about the unlearning set was removed. We apply MIA with logistic regression and SVM \cite{chundawat2023can}. In the unlearning model, the attack probability on \( \mathcal{D}_f \) should be lower, ideally similar to that of the retrained model.
6) \textbf{Execution time}: Finally, we evaluate the time performance of the unlearning method.

Accuracy, perplexity, activation distance, and feature distance can be considered utility metrics, measuring the impact of the unlearning process on the model's utility across different subsets. MIA serves as a privacy metric, assessing the effectiveness of the model's unlearning. Execution time is a computational metric, that evaluates the efficiency of the unlearning training.

\subsubsection{Implementation Details}

For each dataset, we randomly selected samples from the training set as the unlearning set \( \mathcal{D}_f \), with the remaining samples comprising the retained set \( \mathcal{D}_r \). Specifically, for class unlearning, we randomly selected all samples from a particular class as \( \mathcal{D}_f \); for sample unlearning, we randomly selected individual samples as \( \mathcal{D}_f \). We used the same unlearning and retained sets across all unlearning methods. In the training and unlearning processes for image classification tasks, we used the AdamW optimizer with a learning rate of 5e-5 and a batch size of $128$. For the pre-trained ResNet18 model, we trained on CIFAR-10 and Fashion-MNIST for $20$ epochs with $500$ warmup steps to obtain the original model. For ViT, we trained for $10$ epochs with $250$ warmup steps. Text classification and text generation tasks used similar hyperparameters with a batch size of $32$. During the unlearning training, we conducted training for only one epoch with no warmup steps. 
To enhance unlearning, we might train multiple epochs, particularly when the unlearning set is small.

In our main experiments, to comprehensively evaluate our method's feature alignment capabilities, we adopted a consistent strategy of inserting LoRA modules into the linear transformation layers within the model architecture. For ResNet18, LoRA modules were applied to all convolution layers within each residual block. This approach allows the model to adjust residual features across its entire depth. For Transformer-based models, we inserted LoRA modules into the linear layers of the feed-forward network within each attention block, and the final classification head. These components are critical for feature transformation and representation learning \cite{hu2021lora}.

Additionally, we conducted experiments on larger models. Considering computational resources and efficiency, we performed alignment on their classification layers. However, the optimal placement of LoRA modules may vary depending on the task and model. To investigate this, we conducted an ablation study on the impact of layer selection in Section \ref{sec:lora_selection}. In all experiments, unless otherwise specified, the rank $u$ of the LoRA modules was set to $8$, and the scaling factor alpha was set to $32$.

\subsubsection{Baselines}

We compared our unlearning method with the following baseline methods: 
1) \textbf{Original}: The model is trained on the entire training set \( \mathcal{D} \) before any unlearning process is applied. 
2) \textbf{Retrain}: The model is retrained on the retained set \( \mathcal{D}_r \). 
3) \textbf{NegGrad}\cite{golatkar2020eternal}: Fine-tuning the original model on \( \mathcal{D}_f \) using negative gradients. 
4) \textbf{Fine-tune}: Fine-tuning the original model on \( \mathcal{D}_r \).
5) \textbf{BadT}\cite{chundawat2023can}: Using a competent teacher model on \( \mathcal{D}_r \) and an incompetent teacher on \( \mathcal{D}_f \) to transfer knowledge to the student model.
6) \textbf{SCRUB}\cite{kurmanji2024towards}: Using an adversarial training strategy: first, reverse distillation is applied on \(\mathcal{D}_f\) to push the model away from the original model on \(\mathcal{D}_f\), and then fine-tuning is performed on \(\mathcal{D}_r\).

\subsection{Utility Performance}

\subsubsection{Accuracy}

We first compared the accuracy of different unlearning algorithms on image classification and text classification tasks.
Table \ref{tab:cv-acc} shows the results of the sample unlearning and class unlearning task on the CIFAR-10 and Fashion-MNIST datasets using the ResNet18 and ViT models.

\begin{table*}
\caption{Accuracy of image classification tasks in sample unlearning and class unlearning\label{tab:cv-acc}}
\centering
\small 
\setlength{\tabcolsep}{3.65pt} 
\renewcommand{\arraystretch}{1.3} 
\begin{tabular}{cc|ccc|ccc|cccc|cccc}
\bottomrule
\multirow{3}{*}{} & \multirow{3}{*}{} & \multicolumn{6}{c|}{Sample unlearning} & \multicolumn{8}{c}{Class unlearning} \\
\cline{3-16}
 &  & \multicolumn{3}{c|}{CIFAR-10} & \multicolumn{3}{c|}{Fashion-MNIST} & \multicolumn{4}{c|}{CIFAR-10} & \multicolumn{4}{c}{Fashion-MNIST} \\
\cline{3-16}
& & $\mathcal{D}_r$ & $\mathcal{D}_f$ & $\mathcal{D}_t$ & $\mathcal{D}_r$ & $\mathcal{D}_f$ & $\mathcal{D}_t$ & $\mathcal{D}_r$ & $\mathcal{D}_f$ & $\mathcal{D}_{rt}$ & $\mathcal{D}_{ft}$ & $\mathcal{D}_r$ & $\mathcal{D}_f$ & $\mathcal{D}_{rt}$ & $\mathcal{D}_{ft}$ \\
\hline
\multirow{6}{*}{ResNet18} & Original & 100.00&100.00&95.83&100.00&100.00&94.98&100.00&100.00&95.64&97.50&100.00&100.00&95.47&90.60 \\
& Retrain & 100.00&96.88&95.92&100.00&97.66&94.84&100.00&0.00&95.89&0.00&100.00&0.00&96.33&0.00 \\
& NegGrad & 100.00&99.22&95.70&100.00&96.88&94.65&84.54&12.42&80.62&12.20&83.51&9.87&81.87&9.70 \\
& Finetune & 100.00&100.00&95.67&99.99&100.00&94.77&\textbf{100.00}&94.22&95.69&83.20&\textbf{99.99}&94.53&95.76&76.80 \\
& BadT & 100.00&82.81&95.32&99.99&89.84&94.63&99.99&\textbf{0.06}&95.38&0.00&\textbf{99.99}&\textbf{0.02}&95.77&0.00 \\
& SCRUB & 100.00&99.22&95.67&99.99&100.00&94.50&99.97&67.64&\textbf{95.73}&62.40&99.88&6.75&\textbf{96.07}&5.30 \\
& Ours & \textbf{100.00}&\textbf{97.66}&\textbf{95.84}&\textbf{100.00}&\textbf{97.66}&\textbf{94.85}&\textbf{99.99}&\textbf{0.42}&95.52&\textbf{0.00}&\textbf{99.98}&\textbf{0.63}&\textbf{95.94}&\textbf{0.00} \\
\hline
\multirow{6}{*}{ViT} & Original &100.00&100.00&98.70&99.25&99.22&95.06&100.00&100.00&98.64&99.20&99.29&98.90&95.34&92.50 \\
& Retrain & 100.00&99.22&98.74&99.88&96.09&95.35&100.00&0.00&98.64&0.00&99.95&0.00&96.71&0.00 \\
& NegGrad & 99.97&98.44&98.33&99.01&\textbf{96.88}&94.88&37.34&0.00&36.33&0.00&33.87&\textbf{0.00}&33.63&0.00 \\
& Finetune & 99.98&100.00&98.45&99.71&100.00&95.23&99.98&99.94&98.49&98.70&99.80&67.02&\textbf{96.50}&62.60 \\
& BadT & 99.98&98.44&\textbf{98.70}&99.13&99.22&95.11&99.99&15.02&98.56&15.80&99.38&0.58&96.36&0.20 \\
& SCRUB & 99.99&100.00&98.55&99.69&99.22&\textbf{95.29}&99.99&0.00&98.43&0.00&99.72&0.58&96.43&0.20 \\
& Ours & \textbf{100.00}&\textbf{99.22}&\textbf{98.68}&\textbf{99.95}&\textbf{99.22}&94.99&\textbf{100.00}&\textbf{0.00}&\textbf{98.72}&\textbf{0.00}&\textbf{99.95}&\textbf{0.07}&96.41&\textbf{0.00} \\
\toprule
\end{tabular}
\end{table*}

In the sample unlearning experiments, We found that except for the BadT method (with ResNet18), almost all baseline methods achieved high accuracy on \( \mathcal{D}_r \), \( \mathcal{D}_f \), and \( \mathcal{D}_t \). This indicates that a small number of unlearning samples has a minimal impact on the original model's accuracy. 
In the class unlearning experiments, as shown in Table \ref{tab:cv-acc}, except for the fine-tuning method (which almost fails to achieve the unlearning effect), other baseline methods can forget the specified class information. However, we hope that the metrics of the unlearning model on \( \mathcal{D}_r \), \( \mathcal{D}_f \), and \( \mathcal{D}_t \) should be close to those of the retrained model, rather than simply reducing accuracy on the unlearning set.
From Table \ref{tab:cv-acc}, it can be observed that across different datasets and models, our method is either the closest or second closest to the retrained model among nearly all baseline methods. Additionally, we found that in certain experiments, NegGrad, SCRUB, and BadT also achieve results close to those of the retrained model. However, NegGrad can cause a dramatic performance drop on \( \mathcal{D}_t \) in class forgetting tasks, SCRUB’s adversarial training may make forgetting challenging on some datasets, such as class unlearning on CIFAR-10, and BadT experiences rapid performance decline on \( \mathcal{D}_f \) with multiple epochs of unlearning training. As shown in Table \ref{tab:many-epoch-acc}, after two epochs of unlearning, BadT’s accuracy on CIFAR-10 and Fashion-MNIST drops by $42.19\%$ and $25.00\%$, respectively, compared to the retrained model, while our method only shows a difference of $6.26\%$ and $3.91\%$. Table \ref{tab:many-epoch-acc} also shows that even after two epochs, SCRUB maintains accuracy on \( \mathcal{D}_f \) close to that of the original model, indicating lower unlearning efficiency for SCRUB.

\begin{table}
\caption{Training the ResNet18 model for 2 epochs with sample unlearning\label{tab:many-epoch-acc}}
\centering
\small 
\renewcommand{\arraystretch}{1.3} 
\begin{tabular}{c|ccc|ccc}
\bottomrule
\multirow{2}{*}{} & \multicolumn{3}{c|}{CIFAR-10} & \multicolumn{3}{c}{Fashion-MNIST} \\
\cline{2-7}
& $\mathcal{D}_r$ & $\mathcal{D}_f$ & $\mathcal{D}_t$ & $\mathcal{D}_r$ & $\mathcal{D}_f$ & $\mathcal{D}_t$ \\
\hline
Retrain & 100.00&96.88&95.92&100.00&97.66&94.84 \\
BadT & 100.00&54.69&95.26&99.99&72.66&94.66 \\
SCRUB & 100.00&99.22&95.56&99.99&100.00&94.78 \\
Ours & 100.00&90.62&95.81&100.00&93.75&94.75 \\
\toprule
\end{tabular}
\end{table}

When comparing the accuracy of text classification tasks, we used sentiment analysis. 
As shown in Table \ref{tab:text-acc}, in the sample unlearning task, although the NegGrad method closely resembles the retrained model on \( \mathcal{D}_f \), its significant drop in accuracy on \( \mathcal{D}_t \) renders it impractical. The BadT method appears to forget \( \mathcal{D}_f \), but its performance on \( \mathcal{D}_t \) is inferior to our proposed method. In class unlearning tasks, the accuracy of the retrained model can be misleading. For sentiment classification tasks with only two sentiment labels (positive or negative), it is unsuitable to use these as the target unlearning class. Therefore, we chose to consider texts containing specific keywords as a class (in our experiment, we used ``comedy'' as the keyword) and made this class the target unlearning class. In this sentiment classification task, we prefer the model to forget its sentiment bias towards texts containing the keyword. The retrained model (including the fine-tuning method) only retrains on the retained set, failing to achieve this goal. Therefore, in \( \mathcal{D}_f \) and \( \mathcal{D}_{ft} \), we do not use the retrained model as a reference. In this task, we hope the model shows more uncertainty in \( \mathcal{D}_f \) while maximally retaining the original model's performance on \( \mathcal{D}_r \). From this perspective, only our proposed method and the BadT method among the baseline methods meet this objective. In the SCRUB method, the accuracy on \( \mathcal{D}_{rt} \) drops excessively. This is because its adversarial training strategy causes the model to deviate too far from the original model.

\begin{table}
\caption{Accuracy of text classification tasks in sample unlearning and class unlearning\label{tab:text-acc}}
\centering
\small 
\setlength{\tabcolsep}{3.35pt} 
\renewcommand{\arraystretch}{1.3} 
\begin{tabular}{c|ccc|cccc}
\bottomrule
\multirow{2}{*}{} & \multicolumn{3}{c|}{Sample unlearning} & \multicolumn{4}{c}{Class unlearning} \\
\cline{2-8}
& $\mathcal{D}_r$ & $\mathcal{D}_f$ & $\mathcal{D}_t$ & $\mathcal{D}_r$ & $\mathcal{D}_f$ & $\mathcal{D}_{rt}$ & $\mathcal{D}_{ft}$ \\
\hline
Original & 99.96&100.00&93.31&99.95&100.00&93.31&93.36 \\
Retrain & 99.99&96.88&93.17&100.00&-&93.15&-\\
NegGrad & 97.76&\textbf{94.53}&88.48&49.58&\textbf{54.05}&49.47&55.00 \\
Finetune & 99.92&100.00&93.01&99.94&100.00&93.09&93.52\\
BadT & 99.92&\textbf{94.53}&92.92&99.93&55.70&\textbf{93.10}&56.48 \\
SCRUB & 99.99&97.66&\textbf{93.05}&98.86&97.33&91.28&91.46 \\
Ours & \textbf{99.99}&99.21&93.35&\textbf{100.00}&\textbf{54.77}&93.35&\textbf{50.93} \\
\toprule
\end{tabular}
\end{table}

\subsubsection{Perplexity}

For the text generation task, we used perplexity as a performance metric, with lower values indicating better next-word prediction given a context. We conducted experiments using the ELI5-Category dataset. For the sample unlearning task, we randomly selected $128$ samples as the unlearning set. For class unlearning, we chose ``Earth Science'' samples as the unlearning set. As shown in Table \ref{tab:text-perplexity}, in the sample unlearning task, except for NegGrad, our method maintained perplexity on \( \mathcal{D}_r \) and \( \mathcal{D}_t \) similar to other baselines, all close to the retrained model. On \( \mathcal{D}_f \), the perplexity of the retrained model was $42.74$, an increase of $7.78$ compared to the origin model, indicating a decrease in the retrained model's predictive ability on \( \mathcal{D}_f \). Other baseline methods, including ours, showed an increase in perplexity on \( \mathcal{D}_f \) compared to the origin model, but the increase was not significant. This is because, to maintain unlearning efficiency, we only trained all unlearning algorithms on \( \mathcal{D}_f \) for one epoch, while the retrained model was trained for $10$ epochs. For example, for the BadT method, after duplicating \( \mathcal{D}_f \) five times and adding it to the training set, the perplexity on \( \mathcal{D}_f \) changed from $36.66$ to $43.10$, comparable to the retrained model.
In class unlearning, as shown in Table \ref{tab:text-perplexity}, the perplexity of our method is close to that of the retrained model across all metrics except on \( \mathcal{D}_f \). In contrast, the perplexity of NegGrad significantly exceeds the normal range.
Notably, in this experiment, the fine-tuning method was the closest to the retrained model in multiple data subsets. However, as we will see in the subsequent privacy experiments, fine-tuning fails to achieve true unlearning.

\begin{table}
\caption{Perplexity of text generation in sample unlearning and class unlearning\label{tab:text-perplexity}}
\centering
\small 
\setlength{\tabcolsep}{4pt} 
\renewcommand{\arraystretch}{1.3} 
\begin{tabular}{c|ccc|cccc}
\bottomrule
\multirow{2}{*}{} & \multicolumn{3}{c|}{Sample unlearning} & \multicolumn{4}{c}{Class unlearning} \\
\cline{2-8}
& $\mathcal{D}_r$ & $\mathcal{D}_f$ & $\mathcal{D}_t$ & $\mathcal{D}_r$ & $\mathcal{D}_f$ & $\mathcal{D}_{rt}$ & $\mathcal{D}_{ft}$ \\
\hline
Original & 34.46&34.96&43.47&34.45&33.51&43.42&42.24 \\
Retrain & 33.22&42.74&43.46&31.04&41.47&43.42&43.41 \\
NegGrad &43.49&\textbf{47.89}&54.67&*&*&*&* \\
Finetune &\textbf{33.28}&35.61&\textbf{43.34}&\textbf{33.24}&34.30&\textbf{43.28}&42.50 \\
BadT & 35.59&36.66&44.15&36.83&88.00&45.76&84.76 \\
SCRUB & 35.22&37.60&46.00&33.60&\textbf{34.41}&43.27&42.47 \\
Ours & \textbf{34.48}&35.24&\textbf{43.65}&34.51&33.96&43.65&\textbf{42.68} \\
\toprule
\end{tabular}
\end{table}

\subsubsection{Scalability on Large Language Models}

To assess the applicability of our method to modern large language models (LLMs) and safety-critical tasks, we conducted scalability experiments on the WMDP (Weapons of Mass Destruction Proxy) benchmark \cite{li2024wmdp}. WMDP is designed to measure hazardous knowledge in fields such as biosecurity, cybersecurity, and chemical security, serving as a standard for evaluating unlearning in the context of safety.

Our experimental setup involved the Qwen2.5-3B \cite{qwen2.5} model. The ``Original'' model was obtained by fine-tuning the pre-trained Qwen2.5-3B on the WMDP corpus. For the unlearning task, we targeted the `Chemistry' category for removal (class unlearning). The ``Retrain'' model, our gold standard, was obtained by fine-tuning the pre-trained Qwen2.5-3B on the WMDP dataset, but with the `Chemistry' category excluded. We evaluated performance using Perplexity (PPL) and the Membership Inference Attack (MIA).

\begin{table}[h]
\caption{Unlearning performance on Qwen2.5-3B with the WMDP dataset\label{tab:llm-exp}}
\centering
\small
\renewcommand{\arraystretch}{1.3}
\setlength{\tabcolsep}{4.5pt}
\begin{tabular}{c|cccc|c}
\bottomrule
\multirow{2}{*}{Method} & \multicolumn{4}{c|}{Perplexity } & \multirow{2}{*}{MIA} \\
\cline{2-5}
& $\mathcal{D}_r$ & $\mathcal{D}_f$ & $\mathcal{D}_{rt}$ & $\mathcal{D}_{ft}$ & \\
\hline
Original & 2.02 & 2.31 & 2.06 & 2.24 & 1.00 \\
Retrain  & 2.00 & 2.54 & 2.07 & 2.41 & 1.00 \\
\hline
Finetune & 1.93 & 2.38 & 2.00 & 2.28 & 1.00 \\ 
BadT     & 2.06 & 46.18 & 2.11 & 39.53 & 1.00 \\
SCRUB    & 1.93 & 2.37 & 2.00 & 2.28 & 1.00 \\
Ours     & \textbf{2.05} & \textbf{2.37} & \textbf{2.10} & \textbf{2.29} & 1.00 \\
\toprule
\end{tabular}
\end{table}

The results are presented in Table \ref{tab:llm-exp}. Our primary goal is for the unlearning method to emulate the behavior of the Retrain model. As shown in the table, the Retrain model exhibits low perplexity on the retained data ($\mathcal{D}_{r}$ and $\mathcal{D}_{rt}$), similar to the Original model, but shows an increase in perplexity on the unlearning data (rising from $2.31$ to $2.54$ on $\mathcal{D}_f$), indicating it has successfully removed specific knowledge about the target class.

Analyzing the baselines, the BadT method causes catastrophic forgetting, with perplexity on $\mathcal{D}_f$ exploding to $46.18$, rendering the model incoherent on the target domain rather than selectively unlearning. Conversely, Finetune and SCRUB show a decrease in perplexity on $\mathcal{D}_r$ ($1.93$) compared to the Retrain baseline ($2.00$), suggesting potential overfitting to the retained set. Our proposed method strikes a balance, maintaining a utility profile on $\mathcal{D}_r$ ($2.05$) that is highly consistent with the Retrain model ($2.00$) and Original model ($2.02$). On the unlearning set $\mathcal{D}_f$, our method increases perplexity to $2.37$. While this is slightly lower than the Retrain model's $2.54$, it avoids the destructive behavior of BadT and demonstrates a stable adjustment towards the unlearning target.

Regarding the MIA results, we observe that all methods, including the Retrain model, yield a score of $1.00$. This confirms the ``pre-training data contamination'' issue inherent in LLMs: the model's foundational knowledge from pre-training is so robust that standard MIA techniques cannot distinguish between unlearned and original states based on confidence scores alone. Despite this challenge, the perplexity metrics validate that our method effectively aligns the model's behavior with the Retrain baseline without incurring the high computational cost of retraining or the instability of other approximate methods.

\subsubsection{Activation Distance}

We compare our method's activation distance results on image classification and text generation tasks. Activation distance is measured as the mean L2 distance of the prediction probabilities between the unlearning model and the retrained model on different data subsets, as described in \cite{chundawat2023can}. Ideally, the activation distance of the unlearning model should be as small as possible across different data subsets. Tables \ref{tab:cv-ad} presents the activation distance comparisons for different unlearning methods on sample unlearning and class unlearning tasks in image classification. These results indicate that our method generally achieves lower activation distances compared to other baselines, indicating a closer alignment with the retrained model. In sample unlearning, all baselines exhibit similar activation distances on \( \mathcal{D}_r \) and \( \mathcal{D}_t \). However, there is significant variation on \( \mathcal{D}_f \). Specifically, for the ResNet18 model, the activation distance of the BadT on CIFAR-10 and Fashion-MNIST is significantly greater than that of our method. The BadT method's unlearning strategy involves learning from an incompetent teacher, causing the unlearning model to make near-random predictions on \( \mathcal{D}_f \). This ``random guessing'' strategy is a core concept in many unlearning algorithms, which assigns average labels to \( \mathcal{D}_f \). The key difference lies in the handling of intermediate features. BadT's strategy forces the intermediate layers to undergo substantial adjustments to forget \( \mathcal{D}_f \), resulting in significant feature distance, as seen in Table \ref{tab:text-fd}. In contrast, our method constrains the adjustment of intermediate layers, allowing them to extract features similar to those of the retrained model, even when assigning average labels to \( \mathcal{D}_f \). Notably, in the class unlearning activation distance comparison in Table \ref{tab:cv-ad}, BadT achieves comparable results to ours in most cases. This is because, in image classification tasks with balanced class distributions, the prediction probability for a class absent in the training set tends to be close to average, aligning with BadT's ``random guessing'' strategy. Additionally, SCRUB achieves good results in terms of activation distance, indicating the effectiveness of its adversarial training strategy. Nevertheless, in most cases of class unlearning, our method still achieves a comparable activation distance.

\begin{table*}
\caption{Activation distance of image classification tasks in sample unlearning and class unlearning\label{tab:cv-ad}}
\centering
\small 
\renewcommand{\arraystretch}{1.3} 
\begin{tabular}{cc|ccc|ccc|cccc|cccc}
\bottomrule
\multirow{3}{*}{} & \multirow{3}{*}{} & \multicolumn{6}{c|}{Sample unlearning} & \multicolumn{8}{c}{Class unlearning} \\
\cline{3-16}
 &  & \multicolumn{3}{c|}{CIFAR-10} & \multicolumn{3}{c|}{Fashion-MNIST} & \multicolumn{4}{c|}{CIFAR-10} & \multicolumn{4}{c}{Fashion-MNIST} \\
\cline{3-16}
& & $\mathcal{D}_r$ & $\mathcal{D}_f$ & $\mathcal{D}_t$ & $\mathcal{D}_r$ & $\mathcal{D}_f$ & $\mathcal{D}_t$ & $\mathcal{D}_r$ & $\mathcal{D}_f$ & $\mathcal{D}_{rt}$ & $\mathcal{D}_{ft}$ & $\mathcal{D}_r$ & $\mathcal{D}_f$ & $\mathcal{D}_{rt}$ & $\mathcal{D}_{ft}$ \\
\hline
\multirow{4}{*}{ResNet18} & NegGrad & 0.00&0.08&0.02&0.00&0.07&0.03&0.27&1.13&0.30&1.06&0.30&1.02&0.31&1.05 \\
& Finetune &0.00&\textbf{0.07}&0.03&0.00&\textbf{0.04}&0.03&0.00&1.66&0.03&1.46&0.00&1.70&0.03&1.40 \\
& BadT & 0.00&0.92&0.03&0.00&0.78&0.03&0.00&\textbf{0.62}&0.03&\textbf{0.57}&0.00&1.04&0.03&0.89 \\
& SCRUB & 0.00&0.08&0.03&0.00&0.05&0.03&0.00&0.75&0.04&0.69&0.01&\textbf{0.42}&0.03&\textbf{0.40} \\
& Ours & \textbf{0.00}&0.11&\textbf{0.03}&\textbf{0.00}&0.07&\textbf{0.03}&\textbf{0.00}&\textbf{0.63}&\textbf{0.03}&\textbf{0.58}&\textbf{0.00}&\textbf{0.70}&\textbf{0.03}&\textbf{0.60} \\
\hline
\multirow{6}{*}{ViT} & NegGrad &0.00&0.04&0.02&0.01&0.06&0.03&1.11&0.94&1.13&0.96&0.99&\textbf{0.40}&0.99&0.41 \\
& Finetune & 0.00&\textbf{0.01}&0.02&0.00&\textbf{0.05}&0.03&0.00&1.80&0.02&1.79&0.00&1.18&\textbf{0.02}&1.11 \\
& BadT & 0.00&0.04&0.01&0.01&0.05&0.03&0.00&\textbf{0.67}&0.02&0.68&0.01&0.67&0.03&0.62 \\
& SCRUB & 0.00&0.01&0.01&0.00&0.06&0.03&0.00&0.69&0.02&0.69&0.00&0.44&0.03&\textbf{0.40} \\
& Ours & \textbf{0.00}&0.03&\textbf{0.01}&\textbf{0.00}&0.09&\textbf{0.03}&\textbf{0.00}&\textbf{0.69}&\textbf{0.02}&\textbf{0.68}&\textbf{0.00}&0.76&\textbf{0.03}&0.68 \\
\toprule
\end{tabular}
\end{table*}

The comparison results of activation distances for text generation tasks are presented in Table \ref{tab:text-ad}. Given that text generation tasks typically involve a larger number of logits in the final activation layer (e.g., GPT-2 has $50,257$ logits) compared to image classification tasks, this adds more uncertainty to the comparison of activation distances. The experimental results show that the fine-tuning method performs best in reducing activation distance, followed by our proposed method. Table \ref{tab:text-ad} indicates that the NegGrad method significantly impacts model performance, while the BadT method merely worsens the predictions on \( \mathcal{D}_f \) and \( \mathcal{D}_{ft} \), leading to a larger activation distance compared to the retrained model. SCRUB uses an adversarial strategy during training, applying a negative gradient-like update on \( \mathcal{D}_f \) at the outset. This causes the model to quickly deviate from the original model, resulting in a larger activation distance. The core idea behind using the fine-tune method for unlearning is leveraging catastrophic unlearning \cite{french1999catastrophic}. It focuses on learning from the retained data, preserving their features and distribution while reducing performance on the unlearning set. Since the retrained model is also trained on the retained set, their activation distances are similar. However, fine-tuning relies on multiple epochs of training and cannot completely eliminate the model's memory of the unlearning data, potentially retaining some implicit information. In contrast, our proposed method explicitly shifts the features of the unlearning set towards the average features of the retained set, enabling effective unlearning in a shorter time. The original model is trained on a large-scale dataset, making its feature distribution closer to the real statistical distribution. Unlike the BadT method, which aligns the unlearning set with an incompetent random teacher model, causing a distribution shift in \( \mathcal{D}_r \) and \( \mathcal{D}_t \), our method maintains consistency with the original model's probability distribution across all subsets.

\begin{table}
\caption{Activation distance of text generation tasks in sample unlearning and class unlearning\label{tab:text-ad}}
\centering
\small 
\setlength{\tabcolsep}{3pt} 
\renewcommand{\arraystretch}{1.3} 
\begin{tabular}{c|ccc|cccc}
\bottomrule
\multirow{2}{*}{} & \multicolumn{3}{c|}{Sample unlearning} & \multicolumn{4}{c}{Class unlearning} \\
\cline{2-8}
& $\mathcal{D}_r$ & $\mathcal{D}_f$ & $\mathcal{D}_t$ & $\mathcal{D}_r$ & $\mathcal{D}_f$ & $\mathcal{D}_{rt}$ & $\mathcal{D}_{ft}$ \\
\hline
NegGrad & 4.10&4.27&4.07&141.35&142.27&140.35&141.47 \\
Finetune &\textbf{0.43}&\textbf{0.60}&\textbf{0.41}&\textbf{0.44}&\textbf{0.60}&\textbf{0.39}&\textbf{0.42} \\
BadT & 0.72&0.85&0.68&1.24&11.24&1.20&9.17 \\
SCRUB & 1.29&1.43&1.29&1.11&1.26&1.07&1.07 \\
Ours & \textbf{0.60}&\textbf{0.78}&\textbf{0.58}&\textbf{0.71}&\textbf{0.92}&\textbf{0.64}&\textbf{0.71} \\
\toprule
\end{tabular}
\end{table}

\subsubsection{Feature Distance}

We evaluated feature distance on image classification and text generation tasks by measuring the mean absolute relative error between two models’ outputs at specified intermediate layers:
\[
\frac{1}{N} \sum_{x \in \mathcal{D}} \sum_{k=1}^m |x_k^{a} - x_k^{b}|
\]
where \( x_k^a \) and \( x_k^b \) are the features from layer \( k \) of models \( A \) and \( B \), respectively. 

Based on the comparison baseline, we defined two types of feature distances in our experiments:

\textbf{Definition 1}: Distance between the unlearned model and the original model at intermediate layers.

\textbf{Definition 2}: Distance between the unlearned model and the retrained model.

The second definition is more commonly understood and used in many metrics. We include the first definition because comparing feature distances involves numerous intermediate layer outputs, making it relatively complex. Additionally, since the unlearning model typically starts from the original model, while the retrained model might have a completely different initialization and training process, leading to significant changes in intermediate layers, it is natural to use the original model as a baseline for feature distance comparison. In the first definition, we aim for the feature distance of the unlearning model to be as close as possible to the feature distance of the retrained model. In the second definition, we aim for the feature distance to be minimized. In this paper, we conducted experiments for both definitions. Table \ref{tab:cv-fd} presents the comparison results of feature distances based on the first definition for different baseline methods using the ResNet18 model on CIFAR-10 and Fashion-MNIST datasets. Table \ref{tab:cv-fd} represents sample unlearning and class unlearning. It can be seen that in sample unlearning, the feature distance of our proposed method is the closest to that of the retrained model in all cases. This indicates that our proposed method makes the distance between the intermediate layers of the unlearning model and the original model similar to the distance between the intermediate layers of the retrained model and the original model. In the class unlearning experiments shown in Table \ref{tab:cv-fd}, our method achieved feature distances comparable to those of the retrained model in the vast majority of cases.

Table \ref{tab:text-fd} presents the comparison results of feature distances based on the second definition for different baseline methods in the text generation task, with both sample unlearning and class unlearning results displayed in the same table. From the table, it is evident that our proposed method achieves the smallest feature distance in all cases, indicating that our method is closest to the retrained model in the intermediate layers. The BadT method, on the other hand, shows a significantly larger feature distance, indicating that it causes excessive adjustments in the intermediate layers. Our method imposes constraints on the intermediate layers during training, ensuring that the adjustments remain within a reasonable range.

\begin{table*}
\caption{Feature distance of image classification tasks (ResNet18) in sample unlearning and class unlearning\label{tab:cv-fd}}
\centering
\small 
\renewcommand{\arraystretch}{1.3} 
\begin{tabular}{c|ccc|ccc|cccc|cccc}
\bottomrule
\multirow{3}{*}{} & \multicolumn{6}{c|}{Sample unlearning} & \multicolumn{8}{c}{Class unlearning} \\
\cline{2-15}
  & \multicolumn{3}{c|}{CIFAR-10} & \multicolumn{3}{c|}{Fashion-MNIST} & \multicolumn{4}{c|}{CIFAR-10} & \multicolumn{4}{c}{Fashion-MNIST} \\
\cline{2-15}
& $\mathcal{D}_r$ & $\mathcal{D}_f$ & $\mathcal{D}_t$ & $\mathcal{D}_r$ & $\mathcal{D}_f$ & $\mathcal{D}_t$ & $\mathcal{D}_r$ & $\mathcal{D}_f$ & $\mathcal{D}_{rt}$ & $\mathcal{D}_{ft}$ & $\mathcal{D}_r$ & $\mathcal{D}_f$ & $\mathcal{D}_{rt}$ & $\mathcal{D}_{ft}$ \\
\hline
Retrain & 1.65&1.21&1.61&2.30&22.70&2.28&1.92&1.92&1.96&1.88&2.20&2.39&2.23&2.20 \\
NegGrad & 0.22&0.19&0.28&0.42&0.39&0.43&2.76&3.45&\textbf{2.29}&2.42&\textbf{1.65}&\textbf{1.80}&\textbf{1.56}&1.55\\
Finetune &  0.82&0.72&0.78&0.85&0.84&0.84&0.81&0.89&0.81&0.85&0.64&0.65&0.62&1.63\\
BadT & 0.95&0.84&0.97&0.83&0.82&0.80&0.85&1.04&0.83&0.92&0.75&0.77&0.78&\textbf{2.70}\\
SCRUB & 0.85&0.78&0.84&0.66&0.82&0.65&1.42&1.25&1.21&1.23&1.35&1.36&1.32&1.43\\
Ours & \textbf{1.35}&\textbf{1.27}&\textbf{1.33}&\textbf{1.11}&\textbf{1.25}&\textbf{1.13}&\textbf{1.52}&\textbf{1.63}&\textbf{1.49}&\textbf{1.70}&\textbf{1.36}&\textbf{1.37}&\textbf{1.36}&1.37\\
\toprule
\end{tabular}
\end{table*}

\begin{table}
\caption{Feature distance of text generation tasks in sample unlearning and class unlearning\label{tab:text-fd}}
\centering
\small 
\setlength{\tabcolsep}{4pt} 
\renewcommand{\arraystretch}{1.3} 
\begin{tabular}{c|ccc|cccc}
\bottomrule
\multirow{2}{*}{} & \multicolumn{3}{c|}{Sample unlearning} & \multicolumn{4}{c}{Class unlearning} \\
\cline{2-8}
& $\mathcal{D}_r$ & $\mathcal{D}_f$ & $\mathcal{D}_t$ & $\mathcal{D}_r$ & $\mathcal{D}_f$ & $\mathcal{D}_{rt}$ & $\mathcal{D}_{ft}$ \\
\hline
NegGrad &0.55&0.50&0.53&0.73&0.44&0.42&0.61\\
Finetune &2.85&2.88&3.59&3.06&2.34&4.73&2.68\\
BadT &33.27&23.63&20.34&23.50&19.55&22.91&19.31\\
SCRUB &0.86&1.02&0.75&0.80&0.87&1.21& 0.69\\
Ours & \textbf{0.29}&\textbf{0.39}&\textbf{0.29}&\textbf{0.39}&\textbf{0.40}&\textbf{0.38}&\textbf{0.42}\\
\toprule
\end{tabular}
\end{table}

\subsection{Privacy Performance}

We used Membership Inference Attack (MIA), a standard approach in machine unlearning, to evaluate the privacy guarantees of different unlearning models. The core idea of MIA is to train an attack model that attempts to determine whether a specific sample belongs to the training set of the target model. In the context of machine unlearning, MIA is used to verify whether the samples requested for unlearning have been effectively removed from the model. Specifically, we train an attack model, which is a binary classifier, using the output of the unlearning model on the dataset as input to predict the probability that a sample belongs to the training set. We use the test set and the retained set as training samples for the attack model, labeling the test set as $0$ and the retained set as $1$. Instead of directly using the raw output of the unlearning model, we extract abstract features from the outputs to train the attack model, following the approach of \cite{chundawat2023can}, and use SVM and logistic regression as the discriminative models. After training the attack model, we input the abstract features of the unlearning set from the unlearning model to the attack model to obtain the probability of whether the unlearning set was used in the training set. By comparing the performance of the attack model under different unlearning algorithms, we can assess the effectiveness of various methods in protecting privacy. For non-unlearned samples, the attack model will perform better, while for unlearning samples, a lower attack success rate indicates better unlearning effectiveness and stronger privacy protection. Ideally, if the unlearning algorithm is effective, the attack model's accuracy in determining unlearning samples should be close to random guessing (50\%).

However, in the context of machine unlearning, evaluating the performance of MIA requires a delicate balance. For machine unlearning, an extremely low MIA attack success rate might not always be desirable. An excessively low MIA success rate might make the unlearning model more susceptible to MIA attacks \cite{kurmanji2024towards}, as excessive unlearning can cause the model to become unnatural or inconsistent, providing exploitable patterns for attackers. This balance reflects the trade-off between protecting privacy and maintaining model performance: an ideal unlearning algorithm should remove the influence of specific data while preserving the model's overall structure and generalization ability. The retrained model provides an ideal reference point, as it completely excludes the unlearning data, helping us evaluate the effectiveness of the unlearning algorithms.

Depending on our unlearning objectives and tasks, we evaluate MIA performance differently. For sample unlearning, usually aimed at protecting user privacy, our goal is to make the MIA values of the unlearning model as close to the retrained model's MIA values as possible. Based on this, we then strive for the lowest possible MIA attack success rate. For class unlearning, usually aimed at eliminating bias or harmful samples from the model, making it seem as if the model has never encountered such samples, we hope the accuracy on the unlearning class should be as close to zero as possible, and therefore, the MIA value should be as low as possible. However, for text generation tasks, even if the model has not seen a particular class of samples, it should still be able to generate reasonable outputs. Thus, for text generation tasks, we hope the MIA values are as close to the retrained model's MIA values as possible.

We conducted experiments on three main tasks: image classification, text classification, and text generation. For image classification and text classification tasks, we used the entropy of the probabilities output by the unlearning model as the abstract representation of the samples, which were used as input to the attack model, employing SVM as the discriminative model. In the text generation task, due to the high number of logits in the final layer (e.g., $50257$ for GPT-2), a large number of low probability values could introduce additional noise and affect results. Therefore, we selected the probabilities corresponding to the top $10$ tokens and used their entropy as the abstract representation for the attack model's input.

Table \ref{tab:cv-mia} and \ref{tab:text-mia} show the MIA attack success rates for different baseline methods across various tasks. In Table \ref{tab:cv-mia}, for image classification tasks based on the ResNet18 model, our method generally achieved optimal results, with MIA values closest to the retrained model in most cases. On the CIFAR-10 dataset, our method's sample-level MIA value ($0.89$) was very close to the retrained model ($0.86$), and at the class level, our method ($0.00$) was significantly lower than the retrained model ($0.33$), as expected. Table \ref{tab:cv-mia} shows that our method perfectly matched the retrained model at the sample level ($0.98$) for the ViT model on CIFAR-10, while at the class level ($0.00$), it was significantly lower than the retrained model ($0.53$). Results on Fashion-MNIST further confirmed this trend.

\begin{table*}
\caption{MIA values of image classification tasks after sample unlearning and class unlearning\label{tab:cv-mia}}
\centering
\small 
\renewcommand{\arraystretch}{1.3} 
\begin{tabular}{c|cc|cc|cc|cc}
\bottomrule
\multirow{3}{*}{} & \multicolumn{4}{c|}{ResNet18} & \multicolumn{4}{c}{ViT} \\
\cline{2-9}
& \multicolumn{2}{c|}{\makecell{CIFAR-10}} & \multicolumn{2}{c|}{\makecell{FASHION-MNIST}}& \multicolumn{2}{c|}{\makecell{CIFAR-10}}& \multicolumn{2}{c}{\makecell{FASHION-MNIST}} \\
\cline{2-9}
& Sample & Class & Sample & Class & Sample & Class & Sample & Class \\
\hline
Original  &1.00&1.00&1.00&1.00&1.00&1.00&1.00&1.00\\
Retrain  &0.86&0.33&0.95&0.59&0.98&0.53&1.00&0.74\\
NegGrad &0.98&1.00&1.00&1.00&0.99&1.00&\textbf{1.00}&1.00 \\
Finetune &1.00&0.73&1.00&0.82&1.00&1.00&1.00&0.75\\
BadT  &0.06&0.00&0.05&0.00&0.99&0.00&1.00&0.02\\
SCRUB  &0.99&0.62&1.00&0.73&1.00&0.86&1.00&0.24\\
Ours  &\textbf{0.89}&\textbf{0.00}&\textbf{0.93}&\textbf{0.00}&\textbf{0.98}&\textbf{0.00}&\textbf{0.98}&\textbf{0.00} \\
\toprule
\end{tabular}
\end{table*}

In Table \ref{tab:text-mia}, we present the results of MIA on text classification and text generation tasks. For the text classification task, our method performs well at the sample level, achieving near-equivalent performance to the retrained model. At the class level, our method significantly reduces the MIA success rate, aligning with expectations for class unlearning in classification tasks. As previously mentioned, for generation tasks, whether it’s sample unlearning or class unlearning, the output should remain plausible, and thus we aim for MIA values to be as close to the retrained model's MIA values as possible. In the text generation task, we conducted experiments based on DistilGPT2 on the ELI5-Category dataset. As shown in Table \ref{tab:text-mia}, the original model achieved an MIA success rate of $1.00$ at both the sample and class levels, as expected, indicating the model fully remembered the training data. The retrained model achieved an MIA success rate of $0.50$ at both levels, indicating a significant reduction in information leakage. Our method achieved $0.53$ at the sample level and $0.50$ at the class level, very close to the retrained model's performance, which is ideal, indicating that our method effectively protects privacy while maintaining similar performance to retraining. The NegGrad and finetune methods maintained high MIA success rates ($1.00$) similar to the original model, while the BadT method was closest to the retrained model at the sample level but showed excessively low performance at the class level ($0.08$), potentially leading to unreasonable outputs.

\begin{table}
\caption{MIA values for sample unlearning and class unlearning in text classification and text generation tasks\label{tab:text-mia}}
\centering
\small 
\setlength{\tabcolsep}{10pt} 
\renewcommand{\arraystretch}{1.3} 
\begin{tabular}{c|cc|cc}
\bottomrule
\multirow{2}{*}{} & \multicolumn{2}{c|}{DistilBERT} & \multicolumn{2}{c}{DistilGPT2}\\
\cline{2-5}
 & Sample & Class & Sample & Class \\
\hline
Original &1.00&1.00&1.00&1.00\\
Retrain &0.98&0.97&0.50&0.50\\
NegGrad &\textbf{0.98}&1.00&1.00&1.00\\
Finetune &0.99&0.99&1.00&1.00\\
BadT &0.93&\textbf{0.03}&\textbf{0.52}&\textbf{0.08}\\
SCRUB &0.97&0.97&1.00&1.00\\
Ours & \textbf{0.99}&\textbf{0.05}&\textbf{0.53}&\textbf{0.50}\\
\toprule
\end{tabular}
\end{table}

\begin{table*}
\caption{Comparison of training times for different baseline methods (unit: seconds)\label{tab:time}}
\centering
\small 
\setlength{\tabcolsep}{4pt} 
\renewcommand{\arraystretch}{1.3} 
\begin{tabular}{c|cc|cc|cc|cc}
\bottomrule
\multirow{2}{*}{} & \multicolumn{2}{c|}{ResNet18} & \multicolumn{2}{c|}{ViT} & \multicolumn{2}{c|}{DistilBERT} & \multicolumn{2}{c}{DistilGPT2} \\
\cline{2-9}
 & Sample & Class & Sample & Class & Sample & Class& Sample & Class \\
\hline
Retrain &1222.28&1099.67&2065.85&1863.91&2142.00&2003.25&710.46&684.02\\
NegGrad &7.86&12.74&19.39&37.29&57.10&70.71&8.51&10.87\\
Finetune &61.14&55.10&208.39&189.80&213.95&200.10&72.11&69.18\\
BadT &76.35&73.07&334.78&335.80&414.52&414.69&158.22&151.38\\
SCRUB &64.35&65.76&262.02&282.09&311.19&311.28&114.19&112.50\\
Ours &91.71&90.96&297.38&295.89&370.61&369.99&87.42&79.05\\
\toprule
\end{tabular}
\end{table*}

\subsection{Computational Performance}

To evaluate the computational efficiency of our proposed method, we compared the training times of different models and baseline methods across various tasks. Detailed experimental results are presented in Table \ref{tab:time}, with training times measured in seconds. From the table, it can be observed that the retraining method requires the longest time, especially for complex models, highlighting the importance of efficient unlearning methods for complex models. Among all the baselines, the NegGrad method is usually the fastest, as it only requires fine-tuning on the unlearning set. However, as previously discussed, NegGrad can negatively impact model performance. The Finetune method shows decent time efficiency compared to BadT, yet it fails to achieve true unlearning. Our method is generally comparable to BadT in terms of computational efficiency, with similar training times in most cases. Compared to retraining, our total training time across all tasks is reduced by $94\%$. In BadT, two teacher models are used, whereas our method only requires one teacher model in the student-teacher framework, thereby reducing the required computational resources. 

\subsection{Ablation Studies and Robustness Analysis}

\subsubsection{Comparison of Parameter-Efficient Adaptation Methods}
\label{sec:ablation_peft}

To further validate our choice of LoRA as the core component for residual feature alignment, we compared it with other state-of-the-art parameter-efficient fine-tuning (PEFT) methods, specifically PiSSA \cite{meng2024pissa} and DoRA \cite{liu2024dora}. PiSSA initializes the low-rank adapters using principal singular values, while DoRA decomposes weights into magnitude and direction. We also considered LLaMA-Adapter \cite{gao2023llama}, but excluded it from this specific comparison because its architecture (prefixing tokens via attention) is designed for Transformer-based LLMs and is not natively compatible with the CNN backbones (like ResNet) used in this ablation study, nor does it support the zero-latency inference merging required by our efficiency goals.

Table \ref{tab:peft-comparison} presents the results on the ResNet18 model using the CIFAR-10 dataset for class unlearning. We evaluated Accuracy and Feature Distance (FD). For FD, we report two values: the distance relative to the Original model (Definition 1) and relative to the Retrain model (Definition 2).

\begin{table*}[h]
\caption{Comparison of different PEFT methods. Feature Distance is presented as: \textbf{Def 1 (Def 2)}.\label{tab:peft-comparison}}
\centering
\small
\setlength{\tabcolsep}{3pt} 
\renewcommand{\arraystretch}{1.3} 
\begin{tabular}{c|cccc|cccc}
\bottomrule
\multirow{2}{*}{} & \multicolumn{4}{c|}{Accuracy} & \multicolumn{4}{c}{Feature Distance} \\
\cline{2-9}
 & $\mathcal{D}_r$ & $\mathcal{D}_f$ & $\mathcal{D}_{rt}$ & $\mathcal{D}_{ft}$ & $\mathcal{D}_r$ & $\mathcal{D}_f$ & $\mathcal{D}_{rt}$ & $\mathcal{D}_{ft}$ \\
\hline
Retrain & 100.00 & 0.00 & 97.06 & 0.00 & 2.32 (0.00) & 2.52 (0.00) & 2.26 (0.00) & 2.17 (0.00) \\
LoRA    & 99.99 & 0.50 & 96.66 & 0.10 & 1.13 (2.32) & 1.20 (2.40) & 1.11 (2.27) & 1.17 (2.32) \\
PiSSA   & 100.00 & 1.12 & 96.56 & 0.50 & 0.75 (2.04) & 0.81 (2.09) & 0.76 (2.04) & 0.78 (2.05) \\
DoRA    & 99.99 & 0.46 & 96.64 & 0.10 & 3.10 (4.41) & 2.19 (3.42) & 2.49 (3.71) & 2.05 (3.16) \\
\toprule
\end{tabular}
\end{table*}

The results indicate that LoRA achieves the best balance between unlearning efficacy and feature stability. PiSSA exhibits the smallest Feature Distance (e.g., $0.75$ on $\mathcal{D}_r$), suggesting high stability; however, this rigidity compromises its unlearning capability, as evidenced by the higher accuracy on the unlearning set ($\mathcal{D}_f$ Acc of $1.12\%$ vs. LoRA's $0.50\%$). Conversely, DoRA achieves unlearning efficacy comparable to LoRA ($0.46\%$ on $\mathcal{D}_f$), but at the cost of significantly larger Feature Distances (e.g., $3.10$ on $\mathcal{D}_r$). This indicates that DoRA induces excessive perturbation in the feature space, violating our objective of a smooth unlearning transition. Therefore, LoRA stands as a balanced choice, providing robust unlearning performance with controlled feature shifts and minimal computational overhead.

\subsubsection{Impact of LoRA Placement}\label{sec:lora_selection}
An important design choice is which intermediate layers to insert LoRA modules into. Theoretically, a pretrained model's shallow layers capture more general features, while its deep layers capture more task-specific representations. Therefore, one might assume that applying LoRA only to the deeper layers would provide a more effective trade-off between the forgetting effect and computational cost.

To investigate this, we conducted an ablation study on a ResNet18 model for a class-unlearning task on CIFAR-10. We compared four strategies: 1) applying LoRA only to the shallow layers (the front part of the network), 2) only to the middle layers (the middle part of the network), 3) only to the deep layers (the latter part of the network), 4) and to all layers.

\begin{figure}[t]
\centering
\includegraphics[width=0.99\columnwidth]{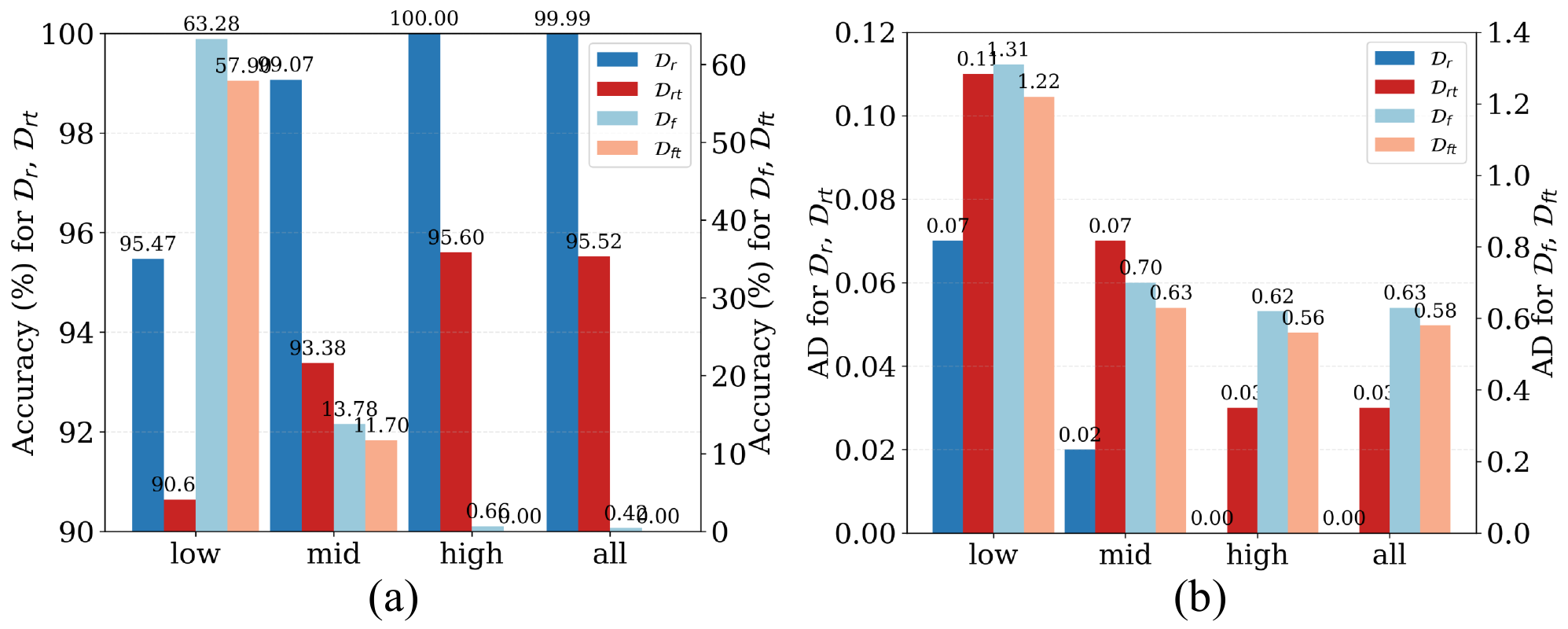}
\caption{The impact of LoRA Placement on accuracy and activation distance. }
\label{fig:layer-selection}
\end{figure}

As shown in Fig.~\ref{fig:layer-selection}, applying LoRA only to the deep layers and to all layers yielded the best performance. These two strategies were comparable in terms of both accuracy and activation distance metrics. Applying LoRA only to the shallow layers was less effective because deep layers are more critical for capturing abstract, class-specific features. This suggests that to achieve efficient and effective unlearning, it is a superior strategy to focus residual feature alignment on the deeper, more task-specific layers.

However, the optimal layer selection may be highly dependent on the specific model architecture and unlearning task, making a single strategy not universally applicable. To ensure the generality and robustness of our proposed method across different tasks and models, we chose to apply LoRA modules to all adaptable layers in our main experiments. In experiments with large models, we inserted LoRA only into the latter half of the network, considering computational resources and efficiency. This approach offers flexibility for different models, allowing them to adjust their feature space as needed during the unlearning process. For maximum effect, we recommend applying it to all layers, while for improved efficiency, targeting the deep layers is a very effective alternative. 

\subsubsection{Impact of $\gamma$ parameter}
The following ablation study evaluates the $\gamma$ parameter, which is a hyperparameter that determines the significance of intermediate layer loss. To assess the impact of $\gamma$, we used the same setup as in previous experiments for sample unlearning and class unlearning, employing the CIFAR-10 dataset and the ResNet18 model. We directly utilized the original and retrained models from the previous experiments, as well as the same samples and classes for unlearning. The unlearning process was carried out using different $\gamma$ values, and the primary metrics evaluated were accuracy and feature distance.

\begin{figure}[t]
\centering
\includegraphics[width=0.99\columnwidth]{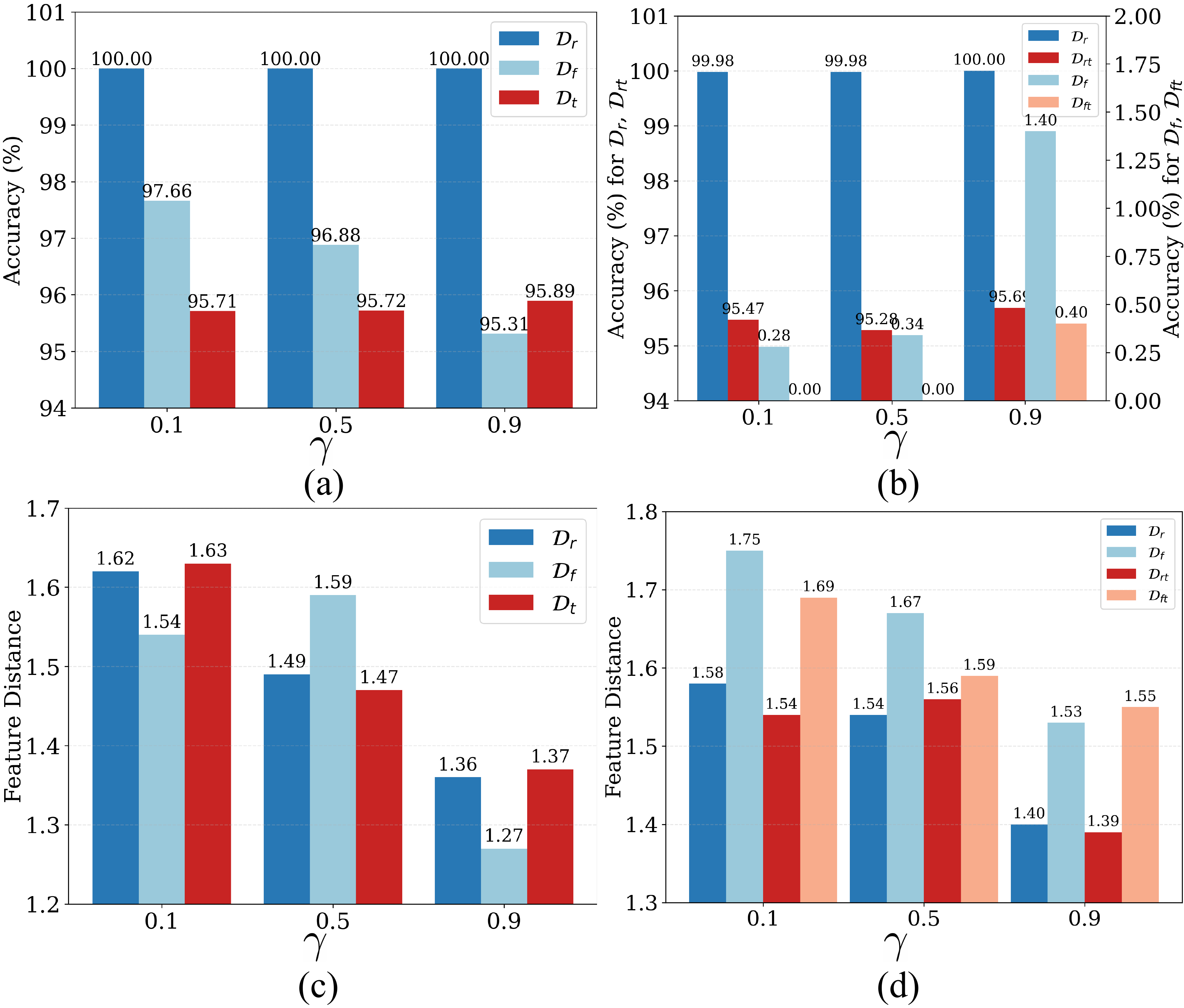}
\caption{The impact of \( \gamma \) on accuracy and feature distance. (a) and (b) show the effect of different \( \gamma \) values on accuracy, while (c) and (d) show the effect of different \( \gamma \) values on feature distance. Among these, (a) and (c) represent sample unlearning, and (b) and (d) represent class unlearning.}
\label{fig:ablation}
\end{figure}

As shown in Fig.~\ref{fig:ablation}, different values of $\gamma$ have minimal impact on the model’s accuracy on $\mathcal{D}_r$ and $\mathcal{D}_t$, while accuracy on $\mathcal{D}_f$ fluctuates within a certain range. Specifically, in sample unlearning, as $\gamma$ increases, $\mathcal{D}_f$ accuracy drops significantly. This is because, with a larger $\gamma$, the model's intermediate features on $\mathcal{D}_f$ are forced to deviate further from their original values and move closer to average features, leading to reduced accuracy on $\mathcal{D}_f$. Interestingly, in class unlearning, this trend is reversed, though the range of fluctuation remains small.

The minimal effect of $\gamma$ on $\mathcal{D}_r$ and $\mathcal{D}_t$ indicates that the intermediate layer loss and the cross-entropy loss in the output layer can substitute for each other to some extent. When $\gamma$ is small, the output layer’s cross-entropy loss dominates; when $\gamma$ is larger, the intermediate layer’s loss takes precedence. However, both losses have a similar influence on model performance, likely because the intermediate layer’s loss reflects the performance of the original pre-trained model, serving as a baseline. Modifying the intermediate layer is equivalent to altering the output layer’s results.

Figure \ref{fig:ablation} shows that $\gamma$ has a linear impact on feature distance: the larger the $\gamma$, the smaller the feature distance, which aligns with our intuition. We observed similar trends in both sample unlearning and class unlearning. In this ablation study, we used the first definition of feature distance, measuring the distance between the model and the original model at the intermediate feature layer. Ideally, the model should retain as much of the original model’s feature extraction capability as possible post-unlearning. The study highlights $\gamma$ as an important factor. Furthermore, in sample unlearning, the feature distance on $\mathcal{D}_{r}$ and $\mathcal{D}_{t}$ tends to be greater than on $\mathcal{D}_{f}$. In class unlearning, however, this trend is reversed, suggesting that the number of forgotten samples has a certain influence on feature distance. Generally, the more samples are forgotten, the greater the feature distance on $\mathcal{D}_{f}$ (or $\mathcal{D}_{ft}$).

\begin{table}
\caption{Robustness to learning rate for class unlearning.\label{tab:robustness}}
\centering
\small 
\renewcommand{\arraystretch}{1.3} 
\begin{tabular}{c|cccc}
\bottomrule
 & $\mathcal{D}_r$ & $\mathcal{D}_f$ & $\mathcal{D}_{rt}$ & $\mathcal{D}_{ft}$ \\
\hline
1e-5 &99.99&0.76&95.47&0.20\\
5e-5 &99.99&0.42&95.52&0.00\\
1e-4 &99.99&0.16&95.59&0.00\\
5e-4 &99.06&0.02&95.01&0.00\\
\toprule
\end{tabular}
\end{table}

\subsubsection{Robustness to Hyperparameters}

To evaluate the robustness of our method, we also analyzed its sensitivity to the learning rate. Using a ResNet18 class unlearning setup on CIFAR-10, we tested learning rates of $1e-5$, $5e-5$, $1e-4$, and $5e-4$. The results in Table \ref{tab:robustness} show that our method is robust. Although a very low learning rate ($1e-5$) performs slightly worse in terms of unlearning (resulting in a higher accuracy on $D_{ft}$), it still achieves effective unlearning (with an accuracy of only 0.2\%). This indicates that our method is robust to changes in the learning rate and does not require extensive hyperparameter tuning to achieve effective unlearning. This robustness stems from the constraints imposed by our feature alignment objective, which provides a regularization effect during the unlearning process. This makes our method more valuable for practical applications.

\section{Conclusion}

This paper presents a novel, general-purpose machine unlearning algorithm that aligns the unlearning model with the pre-trained model at the feature level using residual features from intermediate layers. By integrating LoRA, our method supports both sample-level and class-level unlearning. To demonstrate the versatility of our proposed approach, we conducted extensive experiments across various tasks, including image classification, text classification, and text generation. We evaluated our method using multiple metrics, and the results consistently confirmed its effectiveness and reliability. Future work will focus on achieving unlearning with as few intermediate layer alignments as possible, aiming to balance unlearning performance and computational efficiency.

\section*{Acknowledgments}
This research is supported by the Science and Technology Development Fund (FDCT) of Macau SAR, China (Grant No. fdct 0080/2024/RIA2). This research is also supported by NSFC-FDCT under its Joint Scientific Research Project Fund (Grant No. 0051/2022/AFJ), China \& Macau SAR.

\bibliographystyle{IEEEtran}

\bibliography{references}

\begin{IEEEbiography}[{\includegraphics[width=1in,height=1.25in,clip,keepaspectratio]{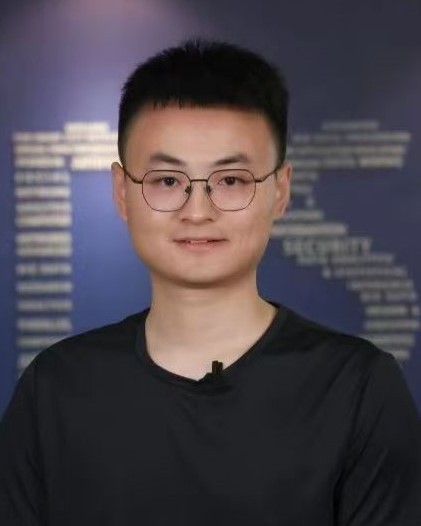}}]{Laiqiao Qin}
is a Ph.D. candidate at City University of Macau, Macau SAR, China. He received his M.Eng. degree in the Faculty of Data Science from City University of Macau. His research interests include AI security and privacy, as well as federated learning.
\end{IEEEbiography}

\begin{IEEEbiography}[{\includegraphics[width=1in,height=1.25in,clip,keepaspectratio]{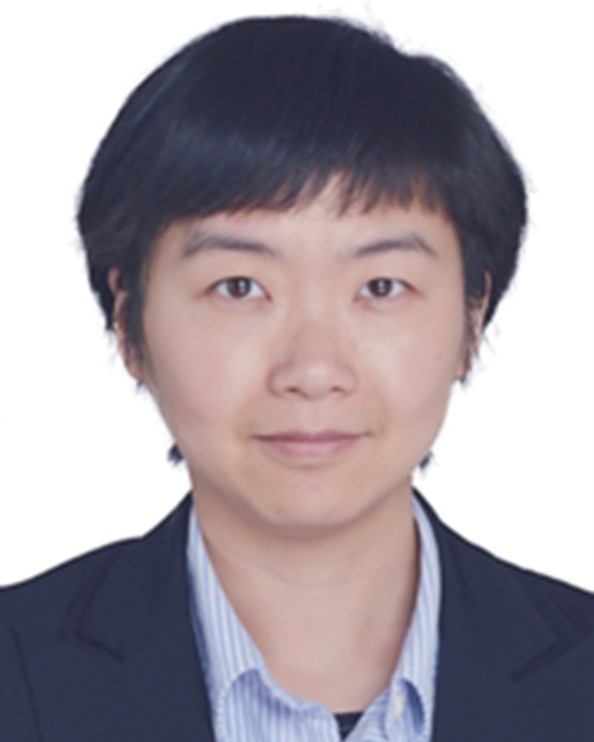}}]{Tianqing Zhu}received her BEng and MEng degrees from Wuhan University, China, in 2000 and 2004, respectively, and a PhD degree from Deakin University in Computer Science, Australia, in 2014. Dr. Tianqing Zhu is currently a professor in the faculty of data science at the City University of Macau. Before that, she was a lecturer at the School of Information Technology, Deakin University, Australia, and an associate professor at the University of Technology Sydney, Australia. Her research interests include privacy-preserving and AI security. 
\end{IEEEbiography}

\begin{IEEEbiography}[{\includegraphics[width=1in,height=1.25in,clip,keepaspectratio]{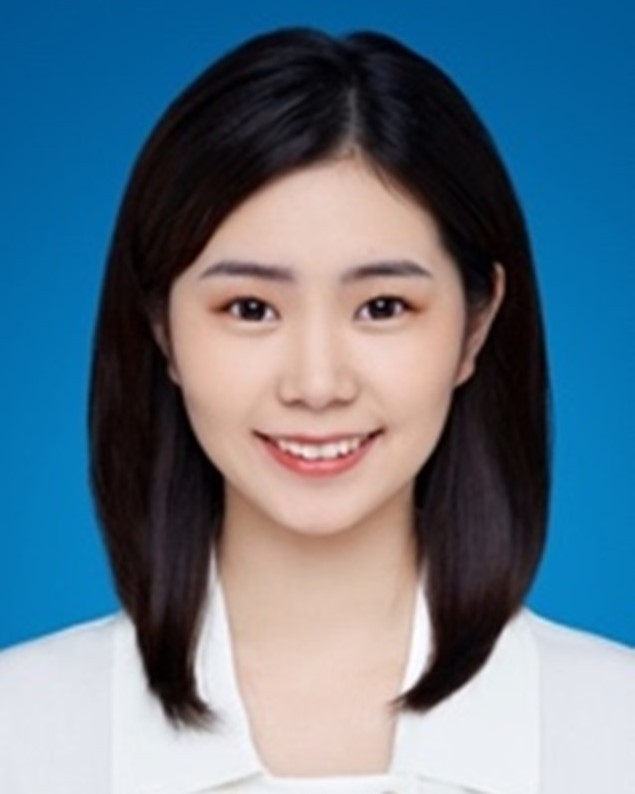}}]{Linlin Wang}
is a Ph.D. candidate at City University of Macau, Macau SAR, China. She received her M.Eng.degree in the Faculty of Data Science from City University of Macau. Her research interests include security and privacy in machine learning, and federated learning.
\end{IEEEbiography}
\begin{IEEEbiography}[{\includegraphics[width=1in,height=1.25in,clip,keepaspectratio]{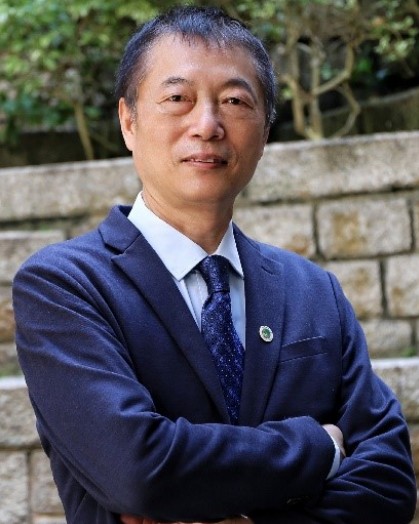}}]{Wanlei Zhou}
is currently the Vice Rector and Dean of Faculty of Data Science, City University of Macau, Macau SAR, China. He received his PhD degree from The Australian National University, Canberra, Australia, in 1991, all in Computer Science and Engineering. He has authored or coauthored more than 400 papers in refereed international journals and refereed international conferences proceedings, including many articles in IEEE transactions and journals. His research interests include security and privacy, parallel and distributed systems, and e-learning.
\end{IEEEbiography}
\vfill
\end{document}